\documentclass[conference]{IEEEtran}
\usepackage[utf8]{inputenc}  

\usepackage[T1]{fontenc}     
\usepackage{lmodern} 
\IEEEoverridecommandlockouts
\usepackage{mathptmx}
\usepackage{url}
\usepackage{cite}
\usepackage{amsmath,amssymb,amsfonts,stmaryrd}
\usepackage{graphicx}
\usepackage{textcomp}
\usepackage{xcolor}
\usepackage{algorithm} 
\usepackage{algpseudocode}
\usepackage{amsthm}
\usepackage{epstopdf}
\usepackage{comment}
\usepackage[caption=false,font=normalsize,labelfont=sf,textfont=sf]{subfig}
\usepackage{array}
\newcolumntype{L}[1]{>{\raggedright\arraybackslash}p{#1}}
\usepackage{makecell}
\usepackage{tikz}
\usetikzlibrary{positioning}
\usepackage{pifont}
\usepackage{array, booktabs, colortbl, xcolor}
\usepackage[table]{xcolor}
\usepackage{booktabs}
\usepackage{colortbl}
\usepackage{placeins}
\usepackage{afterpage}
\usepackage{xurl} 
\usepackage{microtype}
\usepackage{hyperref}

\hypersetup{breaklinks=true}
\definecolor{greencheck}{HTML}{D6F5E3}   
\definecolor{yellowcell}{HTML}{FFF2CC}   
\definecolor{redcell}{HTML}{F4CCCC}      
\definecolor{greycell}{HTML}{EDEDED}     
\definecolor{lightred}{RGB}{255, 204, 204}

\usepackage{graphicx}
\usepackage{svg}
\svgsetup{
  inkscapelatex=false
}

\theoremstyle{definition}

\def\BibTeX{{\rm B\kern-.05em{\sc i\kern-.025em b}\kern-.08em
    T\kern-.1667em\lower.7ex\hbox{E}\kern-.125emX}}
\begin{document}

\title{A Privacy by Design Framework for Large Language Model-Based Applications for Children}
\author{\uppercase{First A. Author}\authorrefmark{1}, \IEEEmembership{Fellow, IEEE},
\uppercase{Second B. Author}\authorrefmark{2}, and Third C. Author,
Jr.\authorrefmark{3},
\IEEEmembership{Member, IEEE}}

\title{A Privacy by Design Framework for Large Language Model-Based Applications for Children\\
{\footnotesize \textsuperscript{}}
}




\author{\IEEEauthorblockN{1\textsuperscript{st} Diana Addae}
\IEEEauthorblockA{\textit{Systems and Computer Engineering} \\
\textit{Carleton University}\\
Ottawa, Canada \\
dianaaddae@cmail.carleton.ca}
\and
\IEEEauthorblockN{2\textsuperscript{nd} Diana Rogachova}
\IEEEauthorblockA{\textit{School of Information Technology} \\
\textit{Carleton University}\\
Ottawa, Canada \\
dianarogachova@cmail.carleton.ca}

\and
\IEEEauthorblockN{3\textsuperscript{rd} Nafiseh Kahani}
\IEEEauthorblockA{\textit{Systems and Computer Engineering} \\
\textit{Carleton University}\\
Ottawa, Canada \\
nafisehkahani@cunet.carleton.ca}

\and
\IEEEauthorblockN{4\textsuperscript{th} Masoud Barati}
\IEEEauthorblockA{\textit{School of Information Technology} \\
\textit{Carleton University}\\
Ottawa, Canada \\
masoudbarati@cunet.carleton.ca}
\and
\IEEEauthorblockN{5\textsuperscript{th} Michael Christensen}
\IEEEauthorblockA{\textit{Department of Law and Legal Studies} \\
\textit{Carleton University}\\
Ottawa, Canada \\
michaelchristensen@cunet.carleton.ca}
\and
\IEEEauthorblockN{6\textsuperscript{th} Chen Zhou}
\IEEEauthorblockA{\textit{School of Information Technology} \\
\textit{Carleton University}\\
Ottawa, Canada \\
chenzhou4@cmail.carleton.ca}
}
\maketitle

\begin{abstract}
Children are increasingly using technologies powered by Artificial Intelligence (AI). However, there are growing concerns about privacy risks, particularly for children. Although existing privacy regulations require companies and organizations to implement protections, doing so can be challenging in practice. To address this challenge, this article proposes a framework based on Privacy-by-Design (PbD), which guides designers and developers to take on a proactive and risk-averse approach to technology design. Our framework includes principles from several privacy regulations, such as the General Data Protection Regulation (GDPR) from the European Union, the Personal Information Protection and Electronic Documents Act (PIPEDA) from Canada, and the Children's Online Privacy Protection Act (COPPA) from the United States. We map these principles to various stages of applications that use Large Language Models (LLMs), including data collection, model training, operational monitoring, and ongoing validation. For each stage, we discuss the operational controls found in the recent academic literature to help AI service providers and developers reduce privacy risks while meeting legal standards. In addition, the framework includes design guidelines for children, drawing from the United Nations Convention on the Rights of the Child (UNCRC), the UK's Age-Appropriate Design Code (AADC), and recent academic research. To demonstrate how this framework can be applied in practice, we present a case study of an LLM-based educational tutor for children under 13. Through our analysis and the case study, we show that by using data protection strategies such as technical and organizational controls and making age-appropriate design decisions throughout the LLM life cycle, we can support the development of AI applications for children that provide privacy protections and comply with legal requirements. 
\end{abstract}


\section{Introduction} \label{sec:introduction}
The growing integration of Artificial Intelligence (AI), including Large Language Models (LLMs) into digital applications for children presents both opportunities and challenges \cite{unicef2021ai,goel2024securing}. On the one hand, educational tools, conversational agents, and LLM-powered storytelling bots can offer children engaging and personalized experiences \cite{zhang2024mathemyths,carmo2024boosting,sun2024storychat}. On the other hand, they also pose considerable privacy risks \cite{goel2024securing}. Recent scholarship and regulatory attention have raised concerns about the large amounts of information collected for training of LLM-based applications and during user interactions, which can involve sensitive and personal data \cite{kibriya2024privacy,Barbera2025_EDPB}. Children are especially vulnerable, as they may not fully understand the risks of exposing private information about themselves in digital environments \cite{gelman2024children, unicef2021ai}. Furthermore, some conversational AI systems are designed to seem friendly and trustworthy, which can encourage children to overshare personal details \cite{kurian2024no}. Additionally, parents often lack sufficient information or guidance to understand how their children’s data are used and processed making it difficult for them to ensure adequate privacy protections \cite{unicef2021ai}. The risks are further exacerbated by LLM-specific vulnerabilities, such as the models’ capacity to memorize and inadvertently reveal sensitive details from training data \cite{carlini2021extracting, carlini2022quantifying}, and by real-world instances where companies providing AI services for children have faced regulatory action after alleged privacy law violations \cite{council2025-buddyai}. Therefore, it is of critical importance to design LLM-based applications with proactive privacy protections that mitigate privacy risks for children. \cite{cohen2019between}. 

National and transnational privacy regulations, including the General Data Protection Regulation (GDPR)\cite{gdpr2016} in the (European Union) EU, the children's Online Privacy Protection Act (COPPA) \cite{coppa2020guidelines} in the United States (US), and ongoing work to strengthen privacy law in Canada \cite{PIPEDA}, agree that children's data deserve special protection in digital contexts \cite{gdpr38,coppa2020guidelines, opcChildrenCode}. For example, GDPR requires stronger safeguards for children's data, as outlined in Article 8, which focuses on parental consent, and Recital 38 \cite{gdprArt8}, which acknowledges the unique privacy needs of children and their limited ability to understand their rights, risks and consequences of data processing \cite{gdpr38}. COPPA is a regulation that applies specifically to online services collecting information from and about children under 13 years of age \cite{coppa2020guidelines}. It requires verifiable parental consent, clear explanations of data use, and limits on data retention among other requirements \cite{coppa2020guidelines}. In Canada, PIPEDA does not have specific rules for protecting children's data. However, the Office of the Privacy Commissioner of Canada (OPC) has recently launched an exploratory consultation to create a Children's Privacy Code \cite{opcChildrenCode}, affirming that children deserve special considerations and protections in digital contexts.

Despite growing recognition of the need to protect children's privacy when designing and deploying AI applications, to our knowledge, there is currently a lack of practical, actionable recommendations for privacy-preserving techniques specifically aimed at Children. Such guidance would help designers and developers create applications that are both legally compliant and secure for children's data. We use “developers”, “designers”, and “engineers” to  refer to companies/organizations, AI service providers, individuals, and policymakers responsible for designing, building and deploying LLM-based systems. While organizations like the United Nations Children's Fund (UNICEF) have issued policy recommendations, such as the Policy Guidance on AI \cite{unicef2021ai}, and the Information Commissioner's Office (ICO) in the United Kingdom has provided guidance through the Age-Appropriate Design Code (AADC) \cite{ico2020aadc}, they offer standards and principles, while focusing less so on practical operational guidance for designers and developers. The academic research on privacy in AI, and LLMs specifically, has also focused on general-purpose systems. In particular, there is increasing research on security and privacy threats in LLMs, including memorization risks and inference attacks \cite{das2025security,chen2025survey,kibriya2024privacy,yao2024survey,carlini2021extracting}. Similarly, there are works on technical risk mitigation strategies, including machine unlearning \cite{ginart2019making,bourtoule2021machine}, differential privacy \cite{abadi2016deep,yu2021differentially}, federated learning \cite{kuang2024federatedscope}, and other approaches (we detail in Section~\ref{sec:related-work}). Another stream of research examined the alignment of LLMs with privacy laws, particularly the GDPR in the European Union \cite{feretzakis2025gdpr,zhang2025right}. Some emerging studies have integrated technical, design, and legal considerations into a framework \cite{al2025framework}. These works provide a foundation for understanding privacy risks, challenges, and mitigation strategies. However, when designing AI applications for children, additional considerations must be taken into account, such as the unique legal protections applicable to children (e.g., COPPA), and the unique vulnerabilities of children, tied to their cognitive, social and developmental capacities \cite{kurian2025developmentally}. To our knowledge, there is no comprehensive framework that adapts regulatory privacy principles and technical controls to the realities of children’s interactions with LLM-based applications. As such, there is a current gap in ensuring that AI applications are both legally compliant and appropriate for children.

To address these, we propose a framework for designing children-focused LLM applications based on the principles of Privacy-by-Design (PbD) \cite{Cavoukian2011PrivacyByDesign}. PbD centers on the idea that service providers should incorporate privacy-preserving mechanisms from the outset and directly within the system's design, business practices, and infrastructures \cite{Cavoukian2012OperationalizingPbD}. In addition, these safeguards should be maintained throughout the lifecycle, not just added later for compliance. The proposed framework maps the regulatory principles of GDPR, COPPA, and PIPEDA, onto the stages of the LLM lifecycle, including data minimization, purpose limitation, meaningful and verifiable consent, security by design, accountability, and user rights. The LLM lifecycle stages involve data collection, model training, operation and monitoring, and continuous validation. For each stage, we provide specific operational controls found in academic literature that translate regulatory principles into technical and organizational actions. These controls aim to mitigate privacy risks and support compliance with regulatory principles. In addition to meeting legal standards, lifecycle mapping enables us to understand how children interact with LLM applications and highlights their unique vulnerabilities and privacy needs. By incorporating privacy protections into every lifecycle stage, the  goal is to guide   designers and developers in designing LLM-based applications that support technical controls, legal compliance, and children-centered design considerations.

The key contributions of the  paper are summarized as follows.
\begin{itemize}
    \item Analyze the privacy violations that arise in LLM-based applications for children, as well as broader risks such as regulatory non-compliance, memorization, profiling and manipulation. 
    \item Propose a PbD-aligned framework that map the core regulatory principles of major privacy regulations into the stages of the LLM lifecycle, including data collection, model training, operation and monitoring, and continuous validation. 
    \item  Connect regulatory expectations and engineering practices by mapping regulatory expectations to practical technical and operational controls that designers and developers can implement. 
    \item Provide an illustrative case study, the educational LLM tutor, along with design recommendations that demonstrate the framework's feasibility in practice and highlight challenges in privacy-preserving LLM deployment. 
\end{itemize}


The remainder of this paper is organized as follows. Sections \ref{sec:background-pbd} and \ref{sec:related-work} introduce the background and related work. Section \ref{sec:threats} discusses cases of privacy violations in LLM-based applications for children, providing justification and motivation for adopting PbD principles in applications designed for children. Section \ref{subsec:regulatory-principles-llms} provides the regulatory principles related to LLM-based applications for children. Section \ref{sec:mapping-principles-llm}  maps the regulatory principles to architectural features throughout the LLM lifecycle, it also discusses an illustrative case study, the educational LLM tutor for children, as a representative application of the framework. In Section \ref{sec:discussion}, we provide an in-depth discussion that summarizes the  findings, reiterates design recommendations, presents the limitations of the proposed approach, assesses both the regulatory suitability and the challenges of implementing the PbD principles, and outlines future work. Section \ref{sec:conclusion} concludes the paper. 

\section{Background}
\label{sec:background-pbd}
In this section, we provide the necessary background on PbD, privacy and data protection laws, and LLMs.
\subsection{Privacy-by-Design (PbD)} 

PbD centers around the idea that simply meeting legal and regulatory compliance does not guarantee adequate privacy protection \cite{HillNotes2021PrivacyByDesign, Cavoukian2011PrivacyByDesign}. Instead, companies and organizations must incorporate privacy as a fundamental aspect of their design processes. This view supports including privacy in organizational practices, information technology systems, and physical structures \cite{HillNotes2021PrivacyByDesign, Cavoukian2011PrivacyByDesign}.

PbD includes seven principles that guide organizations in implementing privacy into the design and development of information systems.
 (1) PbD advocates for proactive measures, which means that organizations should anticipate and reduce privacy risks before they materialize, rather than responding to violations after the fact \cite{Cavoukian2011PrivacyByDesign}.
 (2) Privacy must be the \textit{“default setting”}, with systems automatically protecting privacy, without requiring individuals to turn on privacy protections themselves.
 (3) Organizations must also include privacy protections in the design and setup of their technologies and operations in a way that does not affect system functionality.
 (4) PbD advocates for a\textit{ “positive-sum”} approach, where privacy is seen as compatible with other organizational goals and as part of the overall functionality of a system.
(5) The principle of end-to-end security requires security measures to be implemented to protect personal data throughout its lifecycle, from collection to disposal.
 (6) Organizations should also let stakeholders and independent reviewers evaluate their privacy practices.
 (7) PbD emphasizes the \textit{“respect for user privacy”}. It states that organizations should uphold individual rights and expectations by providing clear notices, establishing strong default protections, and offering options that allow users to control their personal information easily \cite{Cavoukian2011PrivacyByDesign}.

 PbD  emerged during a period of rapid technological development, increased system complexity, and global competition, all of which intensified risks to informational privacy and security \cite{Cavoukian2011PrivacyByDesign}. Since 2022, the widespread development and adoption of AI technologies that collect, infer, and process data at unprecedented scale has made the PbD principles particularly relevant. The application of PbD to AI systems involves incorporating privacy standards throughout the system's life cycle, from data collection through training and deployment to validation \cite{kodakandla2024unified}. In this way, AI service providers ensure they incorporate preventive measures into their system design to reduce privacy risks while supporting compliance with regulatory principles.  

This need for proactive integration of privacy protections reflects broader shifts in how regulatory systems have developed to govern informational environments \cite{cohen2019between, barati2020, barati2023}. Some scholars have noted that traditional administrative law frameworks, built for functions of the industrial era such as rule-making and adjudication, are poorly suited to the dynamics of contemporary data infrastructures and require new governance approaches \cite{cohen2019between}. In the case of AI technologies, which are evolving at an unprecedented pace, the law must adapt rapidly to meet the challenges posed by this new paradigm in information systems.

\subsection{Overview of the Privacy and Data Protection Laws}\label{sec:privacy-laws-overview}
Several legal frameworks exist to govern children's data. The most notable among them are GDPR \cite{gdpr2016}, COPPA \cite{coppa2020guidelines}, PIPEDA \cite{PIPEDA}, the California Consumer Privacy Act (CCPA) \cite{ccpa2018}, the UK’s Age-Appropriate Design Code (AADC) \cite{ico2020aadc}, and UNICEF’s AI for Children guidance \cite{unicef2021ai}. Of these, COPPA, GDPR, PIPEDA, and CCPA are binding laws; the AADC is a statutory code of practice under the UK Data Protection Act 2018 (enforceable but not a standalone statute). Although these frameworks share overlapping objectives such as protecting children's personal data, ensuring meaningful consent, and providing mechanisms that ensure parental oversight, their enforcement mechanisms and technical interpretations vary across jurisdictions. Also, most of these frameworks were developed for traditional systems rather than adaptive AI models.

In this paper, we focus on COPPA, GDPR, and PIPEDA because they represent three mature and complementary regulatory regimes.

\subsubsection{Children’s Online Privacy Protection Act (COPPA)}

COPPA is a United States federal law that regulates how online services collect, use, and disclose personal information from children below the age of 13 \cite{coppa2020guidelines}. It applies to operators of websites, mobile applications, AI-driven platforms, or any online service that is directed at children or knowingly collects data from them.

A set of “Core Privacy Requirements” are articulated in the COPPA Rule (16 C.F.R. Part 312), with supporting guidance from the United States Federal Trade Commission (FTC): 
(i) \textit{providing clear and comprehensive privacy notices} describing data collection, use, and disclosure practices; 
(ii) \textit{obtaining verifiable parental consent} (VPC) before collecting, using, or disclosing a child’s personal information \cite{coppa2020guidelines}; 
(iii) \textit{maintaining the confidentiality, security, and integrity} of children’s data through reasonable procedures; 
(iv) \textit{limiting data collection} to what is reasonably necessary for participation in a game, service, or activity; and  
(v) \textit{allowing parents to review, delete, or refuse} further collection or use of their child’s data \cite{coppa2020guidelines}.

Under Section 312.6, COPPA empowers parents to manage their children’s personal information. It is a requirement for operators to provide accessible mechanisms, such as online dashboards, email verification, or secure communication channels, for parents to exercise these rights \cite{coppa2020guidelines}. However, COPPA does not explicitly include the more advanced rights found in other regulations, such as data portability or challenging automated decision-making, which limits its applicability in modern AI contexts.

The law also specifies acceptable methods for obtaining verifiable parental consent, including signed consent forms, payment card verification, video calls, or government-issued ID checks. Operators are prohibited from conditioning a child’s participation in an activity on the disclosure of more information than is reasonably necessary. COPPA defines “personal information” broadly, covering not only name, address, and email, but also geolocation data, screen names, photos, audio, video, and any persistent identifier used to recognize a user over time and across websites \cite{coppa2020guidelines}.

\subsubsection{General Data Protection Regulation (GDPR)}

The GDPR (Regulation (EU) 2016/679) is a comprehensive legal framework that governs the processing of personal data across the European Union (EU) and European Economic Area (EEA) \cite{gdpr2016}. While GDPR applies to all individuals, Recital 38 explicitly recognizes that children merit specific protection, particularly in digital environments  where they may have little to no knowledge of associated risks and protections. Likewise, GDPR Article 8 demands that the processing of personal data for children under the age of 16 (or 13, where allowed by Member States) be authorized by the holder of parental responsibility \cite{gdpr2016}. These provisions are particularly important for LLM-based applications, which often collect and transform user inputs into internal statistical representations of text that may be reused in opaque inference pipelines.

GDPR is grounded in seven core data protection principles outlined in Article 5(1): (i) \textit{lawfulness, fairness, and transparency}; (ii) \textit{purpose limitation}; (iii) \textit{data minimization}; (iv) \textit{accuracy}; (v) \textit{storage limitation}; (vi) \textit{integrity and confidentiality}; and (vii) \textit{accountability}, as further detailed in Article 5(2) and Article 24. These principles impose strict obligations on AI developers to ensure that personal data is processed fairly, securely, and for clearly defined purposes.

Additionally, GDPR grants individuals a broad set of rights under Articles 12–23, many of which are highly relevant to children’s data. In Article 12, it is stated that privacy notices should be concise, transparent, and presented in a manner easily understood by children. The rights of access, rectification, and erasure (“right to be forgotten”) are also articulated in Articles 15–17. These rights often pose unique challenges for LLM-based applications for children, since user data may be distributed across model parameters and intermediate artifacts (e.g., embeddings, caches) rather than stored as discrete records in traditional databases \cite{miranda2024preserving,carlini2021extracting}. Again, the right to data portability is established in Article 20, while Article 21 provides the right to object to certain processing activities, including profiling.

Another harmful effect of LLM-based applications that may have a detrimental influence on a child's development or well-being arises when decisions are made \emph{solely} through automated processing (including profiling) and produce legal or similarly significant effects. To mitigate these outcomes—particularly critical in LLM-based educational or healthcare applications—additional protective measures are provided in Article 22, subject to specific conditions and exceptions \cite{gdpr2016}. Last but not least, Article 32 further demands technical and organizational measures to ensure data security, addressing risks such as unauthorized access, inference attacks, and adversarial manipulation—threats that are increasingly relevant in the context of generative AI and prompt-based attacks.

Although GDPR contains numerous provisions, we focus on Articles 5, 8, 12–17, 20–22, 24, and 32 in this paper, as they directly address privacy and security challenges in LLM-based applications for children. These articles combined provide a rigorous framework for embedding privacy into the design, training, and deployment of AI systems, ensuring both regulatory compliance and the ethical handling of children's data.

{\footnotesize
\begin{table*}[ht]
  \caption{Key Regulatory Parameters in COPPA, GDPR, and PIPEDA}
  \centering

\begin{tabular}{>{\raggedright\arraybackslash}p{3.0cm} >{\raggedright\arraybackslash}p{3.0cm} >{\raggedright\arraybackslash}p{3.0cm} >{\raggedright\arraybackslash}p{3.0cm}}
  \toprule
  \textbf{\small Aspect} & \textbf{\small COPPA} & \textbf{\small GDPR} & \textbf{\small PIPEDA} \\
  \midrule
  \rowcolor{gray!10}
  \small Age Threshold & \small Under 13 & \small Under 16 (may be lowered to 13 by Member States) & \small No fixed age; capacity-based \vspace{.15cm}\\
  \small Consent Requirement & \small Verifiable parental consent & \small Parental consent required for children & \small Meaningful consent required; maturity assessed \vspace{.15cm}\\
  \rowcolor{gray!10}
  \small Automated Profiling & \small Not explicitly addressed & \small Right not to be subject to automated decision-making & \small Not explicitly addressed \vspace{.15cm}\\
  \small Relevance to LLMs & \small Challenges with verifying consent for dynamic interactions & \small Difficulties with transparency, personalization, and data inference & \small Ambiguity in consent maturity and inferred data handling \\
  \bottomrule
\end{tabular}

  \label{tab:regulatory-parameters}
\end{table*}
}

 \begin{table*}[ht]
  \centering
  \caption{Summary of Privacy Rights Across COPPA, GDPR, and PIPEDA}
\begin{tabular}{p{7.2cm} >{\centering\arraybackslash}p{1.6cm} >{\centering\arraybackslash}p{1.6cm} >{\centering\arraybackslash}p{1.6cm}}
  \toprule
  \textbf{\small Privacy Right} & \textbf{\small COPPA} & \textbf{\small GDPR} & \textbf{\small PIPEDA} \\
  \midrule
  \small Right to Access & \cellcolor{yellow!20}\texttildelow & \cellcolor{green!15}\checkmark & \cellcolor{green!15}\checkmark \\
  \small Right to Rectification & \cellcolor{red!15}\texttimes & \cellcolor{green!15}\checkmark & \cellcolor{green!15}\checkmark \\
  \small Right to Erasure & \cellcolor{red!15}\texttimes & \cellcolor{green!15}\checkmark & \cellcolor{yellow!20}\texttildelow \\
 \small  Right to Restrict Processing & \cellcolor{red!15}\texttimes & \cellcolor{green!15}\checkmark & \cellcolor{yellow!20}\texttildelow \\
  \small Right to Object & \cellcolor{red!15}\texttimes & \cellcolor{green!15}\checkmark & \cellcolor{red!15}\texttimes \\
  \small Right to Data Portability & \cellcolor{red!15}\texttimes & \cellcolor{green!15}\checkmark & \cellcolor{red!15}\texttimes \\
  \small Right to Withdraw Consent & \cellcolor{yellow!20}\texttildelow & \cellcolor{green!15}\checkmark & \cellcolor{green!15}\checkmark \\
  \small Right to Challenge Automated Decisions & \cellcolor{red!15}\texttimes & \cellcolor{green!15}\checkmark & \cellcolor{red!15}\texttimes \\
  \small Right to Lodge Complaints or Seek Redress & \cellcolor{yellow!20}\texttildelow & \cellcolor{green!15}\checkmark & \cellcolor{green!15}\checkmark \\
  \bottomrule
  \end{tabular}
  \begin{center}
  \footnotesize{Note: \checkmark -- Explicitly required; \texttildelow -- Implicit or partial coverage; \texttimes -- Not formally required}
  \end{center}
  
  \label{tab:privacy-rights-summary}
  \end{table*}

\subsubsection{Personal Information Protection and Electronic Documents Act}
Canada’s federal privacy law, PIPEDA, oversees the collection, use, and disclosure of personal information by private-sector organizations during commercial activity operations \cite{PIPEDA}. Unlike COPPA and GDPR, PIPEDA does not establish a fixed age threshold for consent. Instead, it relies on the principle of \textit{meaningful consent}, which must be assessed based on a child’s maturity, cognitive ability, and capacity to understand the implications of data processing \cite{PIPEDA}. The OPC advises that, where a child cannot reasonably provide informed consent, organizations must obtain consent from a parent or guardian. 

Although the law is flexible within its applicable context and accommodates a wide range of digital services, it also introduces ambiguity when it is applied to emerging AI technologies such as LLM-based applications for children, where it is inherently challenging to determine a child’s capacity for consent.

PIPEDA is embodied in ten “Fair Information Principles” enshrined in Schedule~1 of the Act \cite{PIPEDA}. These principles include: (i) \textit{accountability}, requiring organizations to designate a privacy officer and implement internal policies; (ii) \textit{identifying purposes}, ensuring that the reasons for data collection are communicated at or before the time of collection; (iii) \textit{consent}, which must be meaningful and informed; (iv) \textit{limiting collection} to what is necessary for identified purposes; (v) \textit{limiting use, disclosure, and retention}; (vi) \textit{accuracy}; (vii) \textit{safeguards} to protect data; (viii) \textit{openness} about data practices; (ix) \textit{individual access}; and (x) \textit{challenging compliance} through accessible complaint mechanisms. These principles parallel GDPR’s data protection framework but are typically less prescriptive and more flexible in interpretation.

In terms of rights, PIPEDA grants individuals the ability to access their personal data, request corrections for inaccuracies, and receive clear explanations of how their data are being collected, used, and disclosed (Schedule 1, Principles 9 and 10) \cite{PIPEDA}. However, it does not explicitly provide a right to erasure or data portability, which are central features of GDPR. For LLM-based applications for children, this gap poses challenges, particularly when personal data is embedded within model parameters and cannot easily be identified or removed.

In summary, COPPA, GDPR, and PIPEDA form distinct yet overlapping regulatory frameworks designed to safeguard personal data, each with different approaches to consent, age thresholds, and data subject rights. Table~\ref{tab:regulatory-parameters} provides a comparative snapshot of these key parameters, establishing a baseline for understanding how these regulations apply to LLM-based applications for children.

To complement this analysis, Table~\ref{tab:privacy-rights-summary} provides a comparative overview of the privacy rights granted under COPPA, GDPR, and PIPEDA, highlighting areas of alignment as well as gaps that must be addressed when applying these regulations to LLM-based applications for children. This sets the stage for the next subsection, which discusses how these regulatory principles can inform privacy-preserving design strategies in LLM-based applications for children.

\subsection{Large Language Models (LLMs)}\label{sec:LLMs}
This section provides a high-level overview of LLMs. The goal is to provide a broad understanding of the lifecycle stages and the data processing activities involved in the development and operation of LLMs. In addition, this section serves as a technical introduction to the following sections, which will analyze specific privacy principles, how they apply to each lifecycle stage, and the technical controls implemented to support privacy preservation.  

Advances in natural language processing (NLP) have led to the development of LLMs based on the Transformer architecture \cite{liu2023summary}. The Transformer architecture enabled language models to capture long-range dependencies in text \cite{vaswani2017attention}. This design has made LLMs effective in many NLP tasks and enabled the generation of coherent content \cite{naveed2025comprehensive}. Additionally, tuning these models on specific instructions and introducing Reinforcement Learning from Human Feedback (RLHF) have enabled LLMs to achieve strong performance across a wide range of tasks, including machine translation, text summarization, and question answering \cite{liu2023summary}. 

Public interest in these models surged following the release of ChatGPT in late 2022, which accelerated both their deployment and research into their capabilities and limitations \cite{liu2023summary}. More recently, LLMs have been integrated with models that process audio and visual data, resulting in multimodal systems that can understand and generate content across different modalities, such as sound, video, and images \cite{junlin2025large}. Today, LLMs are the foundation for many advanced language-processing technologies, including AI-based virtual agents and multimodal applications \cite{junlin2025large}.

As deep learning-based NLP systems, LLMs follow a multi-stage life-cycle \cite{chen2025survey}. Although implementation details vary depending on the deployment context and supporting infrastructure \cite{Barbera2025_EDPB}, for the purposes of this review, we describe the LLM life-cycle through four main stages: data collection, model training, operation and monitoring, and continuous validation. In the following, we provide a high-level overview of these stages. 

\subsubsection{Data Collection}
Data collection is a foundational phase in the LLM life-cycle. During this phase, raw data is gathered from various sources, including books and digital content \cite{zhao2023survey}. Training corpora typically consist of publicly available datasets, such as BooksCorpus, Common Crawl, and the Colossal Clean Crawled Corpus (C4), as well as additional sources such as academic texts and code repositories \cite{zhao2023survey,gao2020pile}. Generally, LLMs utilize a combination of these sources \cite{naveed2025comprehensive}. Once the data are collected, they undergo pre-processing for training. This pre-processing phase includes quality filtering, data de-duplication, and privacy reduction to eliminate noisy, duplicated, and potentially harmful information. Another necessary task during pre-processing is tokenization, which involves converting raw textual data into sequences of individual tokens \cite{naveed2025comprehensive}. Finally, data scheduling is defined as the process of determining how and when subsets of data will be presented to the model during training \cite{zhao2023survey}.

\subsubsection{Model Training}
The training phase typically comprises several sub-stages: (1) foundation model pre-training, where the model learns general language representations using self-supervised objectives such as next-token prediction; (2) domain adaptation through fine-tuning on task-specific datasets; and (3) alignment, employing techniques like RLHF or preference optimization to align the model's behavior with human values such as honesty, helpfulness, and harmlessness \cite{zhao2023survey,naveed2025comprehensive}.

\subsubsection{Operation and Monitoring}
Once training is completed, LLMs are deployed in production environments to support various applications \cite{IEEE2024_LLM_Lifecycle}. This deployment can occur on cloud services, edge devices, or local servers. After deployment, LLMs interact directly with users to provide real-time outputs \cite{IEEE2024_LLM_Lifecycle}. Therefore, the operation and monitoring phase includes inference, in which the model generates responses based on input prompts and may collect and store user interactions for safety review, debugging, analytics, and future improvements \cite{Barbera2025_EDPB}.

\subsubsection{Continuous Validation}    
After deployment, the continuous validation phase focuses on ongoing monitoring, oversight, and governance of LLM-based applications. Continuous monitoring helps ensure that model outputs remain free of bias or inaccuracies \cite{IEEE2024_LLM_Lifecycle}. In addition, it ensures that model performance remains optimal. This involves tracking metrics such as latency, throughput, and accuracy \cite{IEEE2024_LLM_Lifecycle}. Updates might also include retraining the model on new data, which might come from previous user interactions \cite{Barbera2025_EDPB}. Continuous validation may further include red-teaming and adversarial testing, periodic audits of privacy and security controls, and updating documentation (e.g., model cards and dataset documentation) as the system evolves \cite{gupta2023chatgpt,gebru2021datasheets,mitchell2019model}.

\section{Related Work} \label{sec:related-work}
Building on the preceding discussion of the PbD recommendations, relevant privacy regulations, and the technical foundations of LLMs, this section presents the related work. Existing research has attempted to integrate technical, design, and regulatory perspectives into a unified framework informed by the principles of PbD \cite{al2025framework}. However, the framework proposed by Al Breiki and Mahmoud \cite{al2025framework} is primarily oriented toward general-purpose applications and does not explicitly address child-specific considerations. To address this gap, we adopt their PbD-guided framework as a foundational structure and extend it to incorporate privacy, regulatory, and design considerations tailored to children. This extension is particularly important given growing evidence that children increasingly interact with conversational agents and AI-enabled search tools \cite{ofcom2024-online-nation}.

The remainder of this section reviews three main bodies of related work: (1) technical approaches to privacy protection in LLMs; (2) PbD-guided frameworks for LLM-based applications that integrate regulatory and design considerations; and (3) special vulnerabilities and considerations for children. By synthesizing previous work, we highlight both the progress and gaps, particularly the absence of a unified, comprehensive, and operational framework that integrates regulatory principles, technical controls, and design considerations specific to children. 

\subsection{Technical Privacy Threats and Protections in LLMs}
According to Ofcom’s 2024 Online Nation report, 54\% of children aged 8–15 in the UK used a generative AI tool in the past year \cite{ofcom2024}. These include ChatGPT \footnote{https://chatgpt.com}, Microsoft Copilot \footnote{https://copilot.microsoft.com}, Snapchat MyAI \footnote{https://www.snapchat.com/@myai}, and Google Gemini \cite{ofcom2024}, which are not designed or marketed primarily for children under 13 years of age. Similar patterns have been observed in North America \cite{madden2024dawn}. In response, some companies have begun to introduce child-centred protections. For example, OpenAI has added parental controls that allow parents of teens to link their account with their teen's account and manage safety-related settings \cite{openai2025controls}. Google implemented similar safeguards for its Gemini application. The case of Google is particularly interesting, as the service now allows children under 13 years of age to access the application with parental controls available through Family Link, which provides safeguards such as providing and removing application access and content filtering \cite{jiao2025safe}. However, these measures assume parents are aware whether and how their child is using such applications, which is often not the case \cite{madden2024dawn,eira2025parents}. In addition, there are different parenting approaches, with some parents avoiding the use of restrictive measures to support their children's autonomy \cite{wisniewski2015preventative}.  

As an increasing number of people, including children, have started using LLM-based applications, it is essential to understand the privacy risks associated with these applications. Existing research on privacy in LLMs has found several types of vulnerabilities. Recent surveys classify these vulnerabilities into two main categories based on how attackers can access private information: privacy leakage and privacy attacks \cite{chen2025survey}. Privacy leakage refers to the unintentional release of sensitive information caused by the internal behavior or design of a model system. In contrast, privacy attacks involve intentional actions that take advantage of model or system weaknesses to extract sensitive data \cite{chen2025survey}. 

A common form of privacy leakage occurs when users inadvertently share sensitive information while interacting with LLMs. If such data are logged and later reused for fine-tuning or retraining, it may reappear in future model outputs, posing privacy risks \cite{chen2022backdoor}. Recent work found that LLMs can reveal private information in contexts where humans would not \cite{mireshghallah2023can}. Separately, LLMs have also been known to remember parts of their training data, a phenomenon called memorization \cite{carlini2021extracting}. Under certain conditions, they can reveal this data in their responses \cite{carlini2021extracting,carlini2022quantifying}. This is concerning because memorized examples may contain private details like names, addresses, social security numbers, or snippets of conversations collected from users. Memorization creates an opportunity for adversaries to exploit. In a training-data extraction attack, an attacker recovers memorized training examples verbatim from a model through targeted querying techniques. Similarly, a membership-inference attack determines whether a specific record was part of the training data, while model inversion reconstructs approximate representations of training examples; inversion is similar to data extraction but does not require verbatim recovery \cite{carlini2021extracting}.

To minimize these vulnerabilities, researchers have proposed a range of technical strategies, including but not limited to differential privacy, machine unlearning, and federated learning \cite{kibriya2024privacy,bourtoule2021machine,abadi2016deep,yu2021differentially}. Differential privacy is applied to limit the influence of individual data points on model parameters \cite{behnia2022ew}, while machine unlearning techniques seek to enable post-hoc deletion of sensitive data from trained models \cite{ginart2019making,bourtoule2021machine}. Federated learning allows multiple entities to learn simultaneously without sharing data and sending only model updates to the central server \cite{kuang2024federatedscope,kibriya2024privacy}. In addition to these techniques, researchers have also explored measures such as input and output filtering \cite{shvetsova2025avi}, and safeguards against adversarial behavior \cite{qi2024evaluating}. Transparency artifacts such as dataset documentation \cite{gebru2021datasheets} and model cards \cite{mitchell2019model} have been introduced to make data practices and model limitations more visible to stakeholders. Lastly, organizations can identify and assess risks in LLMs through adversarial testing methods, and offer ways to address these vulnerabilities, thus strengthening accountability \cite{gupta2023chatgpt}. 

\subsection{Privacy-by-Design in AI Systems}
Researchers have extensively explored privacy-preserving techniques in AI; however, comparatively less attention has been given to integrating these technical mechanisms with legal compliance principles. Breiki and Mahmoud \cite{al2025framework} have developed a framework to integrate PbD into generative AI (GenAI) applications. They demonstrate the proposed framework through an AI chatbot in which privacy protections operate in three modes: strict, standard, and personalized. These modes provide different levels of privacy assurance, depending on user preferences, enabling users to have control over their data. The framework also incorporates differential privacy mechanisms to prevent the memorization of personal data and reduce the leakage of sensitive information. Additional mechanisms, such as privacy risk detection tools that warn users when entering PII, and encrypted and access-controlled logs to support regulatory compliance, add another protective layer to the framework. 

Given the novelty of the topic, some of the existing literature is still developing. The emerging work by Sens \cite{sens2025towards} aims to examine the application of the PbD framework in user-based machine learning (ML). The work proposes several methods for embedding PbD into ML-based software, including requirements specification with static verification during implementation and dynamic verification at runtime, as well as data minimization strategies for training ML models. The requirements specification involves creating a domain-specific language for annotating data and system components with regard to their privacy impact, to help developers apply specific measures to more privacy-impacting entities. Static and dynamic privacy verification, as well as data-efficient ML training, focus on measuring the privacy impact of specific data features and their influence on system performance. The goal is to effectively anonymize private data without compromising the utility of the model. Finally, the work proposes a modular architecture for ML-enabled systems that combines these privacy-preserving methods \cite{sens2025towards}.  

The study by Oh et al. \cite{oh2025petlp} introduced the PbD Extract, Transform, Load, and Present (PETLP) compliance framework for embedding legal safeguards directly into data ingestion pipelines. The work specifically deals with social media information, where personal data can be easily ingested by extract, transform, load (ETL) pipelines and poses significant privacy risks. The main proposition of this work is to implement Data Privacy Impact Assessments (DPIAs) prior to data collection and to update them throughout the research lifecycle. In this way, the DPIAs serve as living documents that guide researchers' decisions regarding data; they can help identify risks and assign mitigations early, document security measures, and document ongoing obligations. 

In summary, current research on privacy in LLMs focuses mainly on threats, technical security, privacy protections, and regulatory compliance. To date, one study proposes an integrated framework \cite{al2025framework}. However, there is a lack of frameworks that combine these perspectives specifically for children, leaving gaps in operationalized approaches that simultaneously safeguard data, comply with regulations, and address child-specific design needs. This work builds on this integrated PbD-based framework \cite{al2025framework} by adopting its core structure and operationalizing it within a child-specific context.

\subsection{Specific Considerations for Children}\label{specific-considerations}
In the previous section, we examined threats associated with general-purpose LLM-based applications identified in the literature. Although these systems are primarily designed for adult users, they are increasingly accessed by children. This section focuses on risks specific to children that have been identified in prior research. Research suggests that, due to cognitive, developmental, and emotional differences, children may interact with and interpret LLM-based applications differently from adults, thereby increasing their susceptibility to various harms, including privacy-related risks \cite{kurian2025developmentally,kurian2024no}.
 
Developmental studies summarized by Nomisha \cite{kurian2025developmentally} indicate that young children (ages 2–7) often struggle to distinguish between living beings and machine-based social agents \cite{tanaka2007socialization}. Older children (ages 6–11) also tend to attribute emotional states to non-living entities, for example, describing a voice assistant as “happy” based on its responses \cite{andries2023alexa}. This tendency to ascribe human-like qualities to machines is commonly referred to as anthropomorphism, which can have significant privacy implications \cite{kurian2025developmentally}. During interactions with such systems, children may unintentionally disclose personal information, including their names, ages, and locations. When applications are designed to appear friendly or human-like, children may be further encouraged to share additional personal details \cite{kurian2025developmentally}. Anthropomorphism, defined as the attribution of human traits, emotions, or intentions to non-living entities \cite{darling2015s}, can foster a sense of trust and understanding, thereby increasing the likelihood of information disclosure. Prior research suggests that children may, in some contexts, trust social robots more than humans \cite{abbasi2022can}.

In addition to anthropomorphic design, researchers have identified conversational techniques known as nudging in some chatbot systems \cite{kurian2024no, weidinger2021ethical}. Nudging often takes the form of follow-up prompts, such as “tell me more,” intended to sustain engagement. However, in interactions with children, such prompts may inadvertently encourage the disclosure of information that children might not otherwise share \cite{kurian2024no}.

Recent work has proposed policy and design recommendations aimed at mitigating the effects of anthropomorphism and nudging \cite{kurian2025developmentally, akbulut2024all}, particularly where these techniques may exploit children’s developmental vulnerabilities and undermine privacy. These considerations are increasingly important as applications designed for children, including interactive tutors, storytelling agents, and gaming applications more frequently incorporate LLMs capable of engaging in open-ended conversations.

\subsection{Summary}
The review of existing literature suggests that current research on LLM vulnerabilities, legal compliance and policy recommendations for child-specific design exists in parallel rather than in combination. This highlights the need for a comprehensive framework that integrates technical protections, legal compliance, and child-oriented design recommendations. In practice, this would involve implementing current policy recommendations and research regarding children and AI, such as designing age-appropriate interfaces and establishing parental consent workflows. Additionally, incorporating technical security measures like differential privacy and machine unlearning. By combining technology-specific safeguards with child-centred approaches, we can achieve more holistic protections for children's privacy. Our contribution to the current body of work is to merge these established practices into a  framework that ensures service providers of child-oriented AI technologies (1) comply with privacy regulations, (2) prevent privacy violations by implementing robust technical protections from the outset, (3) and uphold ethical principles that respect children's best interests and rights. 

\section{Privacy Violations in LLM-based Applications for Children}
\label{sec:threats}

To motivate and emphasize the need to integrate privacy protections into the design of child-focused LLM-based technologies, we offer an overview of recent privacy \emph{violations} and \emph{incidents} resulting from noncompliance \emph{or disputed practices} with applicable privacy regulations across various AI applications used by or developed for children. Although not all systems discussed in this section involve LLMs, they demonstrate the harms that can result from inadequate privacy protections, particularly for vulnerable populations such as children. \emph{Further, several cases involve data flows that closely mirror LLM pipelines, including large-scale data collection for training, logging and retention of user interactions, and third-party processing.} We also provide a brief explanation of what compliance might look like according to the PbD approach. \emph{(Under GDPR, consent is one lawful basis; where consent is relied upon for information society services offered directly to a child, Article 8 adds parental authorization requirements \cite{gdpr2016}.)}

The violations we review include inadequate consent mechanisms \cite{council2025-buddyai}, collection of children’s photographs for model training \cite{verge2024-meta,aiaaic2023-amazon-voice}, design flaws exposing user profiles \cite{characterai2024-incident}, and large-scale surveillance in educational environments \cite{o2025school}. Across these examples, safeguards are often reactive and do not address the unique vulnerabilities of children, such as their greater likelihood of disclosing personal information, the sensitivity of children’s data, and their susceptibility to exploitation. Furthermore, many do not take into account the special legal protections applicable to children, such as the requirement for parental consent and limits on data collection and sharing \cite{coppa2020guidelines}.  

\subsection{Data Collection without Adequate Consent Mechanisms}

Despite the centrality of consent in the major privacy regulations, some providers do not implement adequate mechanisms to obtain, verify, and enforce parental consent. For example, Buddy AI \cite{farkhodovich2024exploring}, a popular AI-based application for children under 12 years old, was found by the Children’s Advertising Review Unit (CARU) to have violated COPPA and the CARU Privacy Guidelines by collecting children’s personal information \emph{(as defined under COPPA)} without providing direct notice, posting visible privacy policies, or obtaining verifiable parental consent \cite{council2025-buddyai}. Although the company eventually updated its practices, the changes were reactive and made only after regulatory scrutiny. This reactive posture contradicts the principles of PbD, which require companies to implement privacy protection mechanisms, such as consent, as default design components \cite{Cavoukian2011PrivacyByDesign}. In LLM-based applications designed for children, compliance would look like providing detailed parental notices explaining data collection and use, a mechanism to provide and revoke consent at any time (i.e., dynamic consent, which can be granted, reviewed, and revoked over time), accessible privacy policies, and clear opt-out procedures. 

\subsection{Unauthorized Training Data Collection}

Training LLMs often involves scraping large amounts of online content, which creates a risk of collecting data about children without their or their parents’ knowledge or consent. For example, Meta scraped public images and posts to train its generative AI models, including photos of children taken from adult-owned accounts \cite{taylor2024meta,malwarebytes2024-meta}. Similarly, the LAION-5B dataset \cite{LAION5B} contained images of children from Brazil and Australia, many of which were labelled with names, ages, and locations \cite{aiaaic2023-amazon-voice,guardian2024-meta-australia}. Researchers at the Stanford Internet Observatory also identified child sexual abuse material (CSAM) within the LAION-5B dataset \cite{stanford2023-laion-report}. Following this discovery, LAION reported that it was working to remove the harmful content \cite{LAION5B}.
These cases reveal significant shortcomings in data collection practices for AI systems. In particular, companies that scrape images from the internet may also collect content from social media platforms for training purposes, even when consent for such use has not been provided. A PbD approach would require upstream safeguards, such as obtaining explicit consent to use data for model training \emph{(e.g., via opt-in licensing or purpose-limited agreements where feasible)} or implementing mechanisms to detect and exclude children’s images from training datasets. \emph{However, general privacy-policy terms alone may be insufficient to constitute explicit consent, and should not be treated as a substitute for transparent, purpose-limited sourcing.} Existing mitigation techniques include child face detection tools, although it is still an active area of research \cite{caetano2025neglected,kireev2025manually}. For example, the Stanford study researchers used perceptual hash-based detection, cryptographic hash-based detection, and k-nearest neighbour analysis to detect children's images \cite{stanford2023-laion-report}. \emph{These automated methods also have limitations (e.g., false positives/negatives and potential bias) and therefore require governance and evaluation when used in practice.} Importantly, these measures should be applied before data are incorporated into AI training. 

\subsection{Accidental Exposure and Public Disclosure of Sensitive Information}

Poor security designs can result in accidental disclosure of sensitive information. In December 2024, Character.AI briefly exposed user account information, which “depending on the user, could include a username, name, bio, Characters, homepage, personas, voices or chats” for approximately 10 minutes \cite{characterai2024-incident}. The company minimized the impact, noting that the breach affected fewer than 0.01\% of the accounts \cite{characterai2024-incident}. However, given the \emph{popularity of generative AI-based applications among children aged 8–15 in the UK} \cite{ofcom2024}, and reports of children using this platform for therapy or emotional support \cite{yu2025exploring}, there is a high risk that such breaches can have serious privacy implications for children.  

Character.AI's broader data-collection practices exacerbate these risks. The platform collects a significant amount of personal data, including conversations, uploaded media files, and voice recordings, and shares some information with third-party advertising and analytics vendors \cite{characterai2024-privacy-policy}. Additionally, their privacy policy indicates the company engages in profiling of users \emph{based on} its collection of user interests and “preferences based on account settings or feedback on the Services” \cite{characterai2024-privacy-policy}. Embedding preventive safeguards, such as restricted collection, minimal retention, and strong encryption, would respect privacy of users, and protect sensitive data \emph{in the} event of a breach.

\subsection{Surveillance in Educational Settings and Data Breaches}

Large-scale AI surveillance in educational settings can significantly undermine children’s privacy. For example, Gaggle Safety Management, a system used to monitor about six million students for indicators of cyberbullying, self-harm, violence, and other behavioral or emotional risks, had recently faced a privacy breach \cite{GaggleSafetyManagement, detroitnews_2025}. In 2025, \emph{Vancouver Public Schools (Washington State, U.S.)} inadvertently exposed nearly 3,500 sensitive student records, including poems, essays, and AI chat transcripts collected by Gaggle, through unprotected links \cite{detroitnews_2025, ap2025gaggle}. This case is an example of the dangers of large-scale surveillance and weak security controls. The absence of basic safeguards, such as password protection, encryption, and access controls, led to a significant privacy breach involving highly sensitive personal information of schoolchildren. Some of this data involved chats about being victims of violence, threats of suicide, and questions about sexuality, the disclosure of which could potentially impact the well-being of these children and have long-term consequences \cite{detroitnews_2025, ap2025gaggle}. Recent research also highlights educators’ concerns about the intrusive nature of monitoring tools and their implications for student trust and autonomy \cite{garcia2025keeping}.

The cases presented in this section illustrate gaps in regulatory compliance among many AI service providers, particularly those that market their technologies to children, those whose products are used by children, or those involving children. A common theme across these cases is that companies often implement privacy protections only after violations occur, treating these safeguards as fixes rather than as integral components of the design process. As various LLM-powered applications, such as educational tutors, companions, and storytelling bots, become more common in children’s daily lives, it is critical to incorporate the principles of PbD into the development, deployment, and monitoring of these systems to ensure adequate privacy protections for sensitive children's data.

\section{Regulatory Principles Relevant to LLM-Based Applications for Children} 
\label{subsec:regulatory-principles-llms}

Building on the comparative analyses of regulatory principles and privacy rights discussed in Section \ref{sec:privacy-laws-overview}, along with the privacy violations outlined in Section \ref{sec:threats}, we identify seven overarching principles: \textit{Data Minimization}, \textit{Purpose Limitation}, \textit{Transparency and Explainability}, \textit{User Rights}, \textit{Accountability}, \textit{Security by Design}, and \textit{Meaningful Consent Mechanisms}. Table~\ref{tab:privacy-principles-summary} summarizes how these principles are articulated in COPPA, GDPR, and PIPEDA. Collectively, these principles form the foundation of our framework, which provides guidance on proactively embedding privacy principles throughout the LLM lifecycle --from data collection and training to deployment and real-time interactions. The following paragraphs elaborate on each principle in the context of LLM-based applications for children and offer a brief overview of controls presented in recent academic literature that can be implemented in practice to help organizations align their practices with the legal requirements for privacy protection. These strategies are explored in more detail and in relation to each of the LLM lifecycle stages in Section \ref{sec:mapping-principles-llm}. 

\begin{table*}[t]
\centering
\caption{Summary of Privacy Principles Across COPPA, GDPR, and PIPEDA}
\label{tab:privacy-principles-summary}
\renewcommand{\arraystretch}{1.4}
\setlength{\tabcolsep}{10pt}
\rowcolors{2}{gray!5}{white}

\begin{tabular}{p{7.0cm} >{\centering\arraybackslash}m{2cm} >{\centering\arraybackslash}m{2cm} >{\centering\arraybackslash}m{2cm}}
\toprule
\textbf{Privacy Principle} & \textbf{COPPA} & \textbf{GDPR} & \textbf{PIPEDA} \\
\midrule
\rowcolor{gray!10}
Data Minimization         & \cellcolor{yellow!20}\texttildelow & \cellcolor{green!15}\checkmark & \cellcolor{green!15}\checkmark \\

Purpose Limitation        & \cellcolor{yellow!20}\texttildelow & \cellcolor{green!15}\checkmark & \cellcolor{green!15}\checkmark \\

Transparency and Explainability & \cellcolor{green!15}\checkmark & \cellcolor{green!15}\checkmark & \cellcolor{yellow!20}\texttildelow \\

User Rights               & \cellcolor{yellow!20}\texttildelow & \cellcolor{green!15}\checkmark & \cellcolor{green!15}\checkmark \\

Accountability            & \cellcolor{yellow!20}\texttildelow & \cellcolor{green!15}\checkmark & \cellcolor{green!15}\checkmark \\

Security by Design        & \cellcolor{green!15}\checkmark & \cellcolor{green!15}\checkmark & \cellcolor{green!15}\checkmark \\

Meaningful Consent Mechanisms & \cellcolor{green!15}\checkmark & \cellcolor{green!15}\checkmark & \cellcolor{green!15}\checkmark \\
\bottomrule
\end{tabular}

\vspace{1mm}
\begin{center}
\small\textbf{Note:} \checkmark — Explicitly required; \texttildelow — Implicit or partial coverage; $\times$ — Not formally required
\end{center}
\end{table*}

\textit{Data Minimization} is explicitly required under GDPR Article 5(1)(c) and PIPEDA Principle 4. Under GDPR, personal data must be “adequate, relevant, and limited to what is necessary” for the specified purpose for which it is processed \cite{gdpr2016}. PIPEDA similarly requires organizations to collect only the information needed for clearly defined purposes, be honest about the reasons for collecting information, and do so by lawful means \cite{PIPEDA}. Although COPPA does not state a standalone data-minimization clause, the principle is implicit and partially mentioned throughout several sections of the rule. 

To support compliance with the principle of data minimization in practice, service providers must enforce the collection of only data items relevant to the purpose, while excluding unrelated ones. For instance, when users are interacting with an application for educational purposes, the system might enforce collection of data from user prompts that are only relevant to that context. Live and dynamic interactions between children and LLM-based applications present unique risks, as these open-ended conversations can lead to unintended disclosures of sensitive or identifiable information. Service providers should therefore pay particular attention to ensure that the system does not collect sensitive information from children unintentionally. Technical solutions exist to support this; for example, the method proposed in \cite{ngong2025protecting} filters user prompts in real time to reformulate and redact potentially sensitive and unrelated information. This approach is discussed in more detail in Section~\ref{sec:mapping-principles-llm}. Service providers should apply such data minimization controls at every point where new information may be collected, including user registration, ongoing interactions, and feature upgrades that involve processing new types of data.

\textit{Purpose Limitation}, as outlined in GDPR Article 5(1)(b) and PIPEDA Principle 5 (“Limiting use, disclosure, and retention”), restricts the use of personal data to the purposes stated at the time of collection and prohibits subsequent re-use for unrelated purposes. In LLM-based applications for children, this means that data collected for one purpose, such as providing educational or interactive services, cannot later be reused for profiling, marketing, or analytics. Service providers should maintain records of the original collection purposes and enforce them whenever the system performs operations on the data, supporting compliance at all stages of the LLM lifecycle. One technical approach proposed in \cite{tokas2020formal} involves tagging personal data with subject–purpose pairs and allowing the runtime system to verify that all data access operations comply with the consented policies. Such mechanisms can help operationalize purpose limitation along with other principles in practice.

\textit{Transparency}, required by Articles 12-13 of GDPR, COPPA Section 312.4, and PIPEDA Principle 8, obliges service providers to communicate clearly about their data practices, particularly to parents and guardians. We use the terms “Transparency and Explainability” to describe the obligation to provide accessible explanations of how personal data are collected, stored, used, and shared, as specified across these regulatory frameworks. Under Section 312.4, service providers must ensure that parents and guardians receive clear notices about the collection and use of their children’s personal data before verifiable parental consent is obtained \cite{coppa2020guidelines}. Such notices must be presented in an easily understandable format, provide complete and accurate information, and avoid unrelated, misleading, or confusing content \cite{coppa2020guidelines}. Service providers can operationalize this by including dashboards for parents and guardians into the design of their applications; persistent privacy policy toggle that can be accessed at any time by parents and guardians to review data collection and use practices; enabling parents to ask questions and seek clarification on data practices in real-time through easily accessible AI service agent or similar service. In LLM-based applications, the transparency might also include explanations on how the system processes data and arrives at its responses. Such explanations can be provided again, in a concise and clear manner that allows parents and guardians to understand, and the underlying mechanisms. Additionally, recent research highlighted that parents and guardians struggle to understand the full range of risks that generative AI technologies present to children \cite{yu2025exploring}. As a consequence, these researchers discussed that service providers include “expert-identified risk taxonomy” which supplies parents with detailed information on the risks posed by a technology \cite{yu2025exploring}. This type of disclosure can be added to parental notices to help them better understand the risks involved with regard to the tool processing private information and make well-informed decisions.  

Although COPPA focuses primarily on providing information to parents and guardians, GDPR Article 12 additionally requires that information be provided to a child in a clear and accessible manner \cite{gdpr2016}. This means meeting children where they are at, providing information in a way that children can understand, and allowing children to meaningfully participate in decision-making. Recent literature offers multiple ways to make lengthy legal explanations more intelligible and participatory for children, including the implementation of visual or animated explanations \cite{ milkaite2020child}. In addition, research has increasingly looked at the mental models of AI in children that can support the development of explainable AI for this demographic \cite{dangol2025children}.

\textit{Meaningful Consent Mechanisms} are central to COPPA Rule Section 312.5, GDPR Articles 6 and 8, and PIPEDA Principle 3. In this paper, we use the term “meaningful consent mechanisms” as an umbrella concept encompassing verifiable parental consent under COPPA, parental authorization under GDPR Article 8, and meaningful consent under PIPEDA. In LLM-based applications for children, parental consent interfaces and notices must be presented any time before new information from a child is collected. For example, when first registering for the app, when assessing its main functionality, and when a child is requesting other services or upgrades that involve processing new types of information (e.g., voice). Service providers must also, using reasonable methods, confirm that the person who provides the consent is the child’s parent or legal guardian (i.e., through identity card verification, knowledge-based questions) \cite{coppa2020guidelines}. Additionally, after initial authorization, providers must offer intuitive interfaces that allow guardians to monitor, modify, or revoke permissions in real time, providing dynamic consent functionality. Dynamic consent has been explored in literature \cite{tokas2020formal}, which could be adapted for systems incorporating LLMs.

\textit{User Rights}, such as access, correction and erasure (GDPR Articles 15–17, PIPEDA Principle 9 and COPPA Rule Section 312.6), are particularly challenging to uphold in trained LLMs due to the data stored as distributed representations instead of raw records \cite{feretzakis2024privacy}. Although there are methods such as machine learning that help remove sensitive information learned from trained models, issues remain \cite{feretzakis2024privacy, blanco2025digital}. They are related to the guaranty of forgetting, as well as the challenge of accessing and deleting explicit data that have been incorporated into the model parameters; as well as implicit sensitive data that do not directly identify individuals, but can reveal the identity when used in combination with other information \cite{blanco2025digital}. To avoid these challenges, developers can implement tools that locate and manage stored interaction records and any user-linked data and remove private or sensitive data before it is used in AI fine-tuning or enhancement pipelines. Modern foundational model pre-processing pipelines already use training data filtering for PII detection and removal \cite{naveed2025comprehensive}. Service providers can also provide clear notices, pathways and interfaces that allow parents or guardians to exercise these rights, such as requesting deletion of their child’s data, including conversation records, voice recordings, images, and other media, in a clear and actionable manner.

\textit{Security by Design}, mandated by GDPR Article 32, PIPEDA Principle 7, and COPPA Rule Section 312.8, emphasizes protecting personal data against security threats. In LLM-based applications, multiple vulnerabilities have been identified that can occur at different stages of LLM development and deployment. They include  adversarial attacks such as data poisoning, backdoors, gradient leakage and membership inference attacks \cite{das2025security, yao2024survey}. Jailbreaking and prompt injection \cite{das2025security, yao2024survey}. Correspondingly, a range of defense mechanisms have been explored to improve model security. Recent studies and surveys highlight mitigation strategies including pruning \cite{chapagain2025pruning}, differential privacy \cite{behnia2022ew}, identification of malicious prompts, and the use of guardrails to prevent the generation of harmful or unsafe content \cite{yao2024survey}. The established techniques such as differential privacy can introduce noise during training or fine-tuning, significantly reducing the risk of sensitive data extraction from trained models \cite{behnia2022ew}. Service providers should ensure that their models, along with the associated software and hardware, are effectively secured to protect against security attacks and unintended breaches.

\textit{Accountability}, as emphasized in Article 5(2) of GDPR and PIPEDA Principle 1, requires organizations to demonstrate compliance with the privacy obligations outlined above. Under PIPEDA, this principle specifically requires organizations to assume the responsibility for compliance, typically designating 

\begin{figure*}[!t]
\centering
\includegraphics[width=0.75\textwidth]{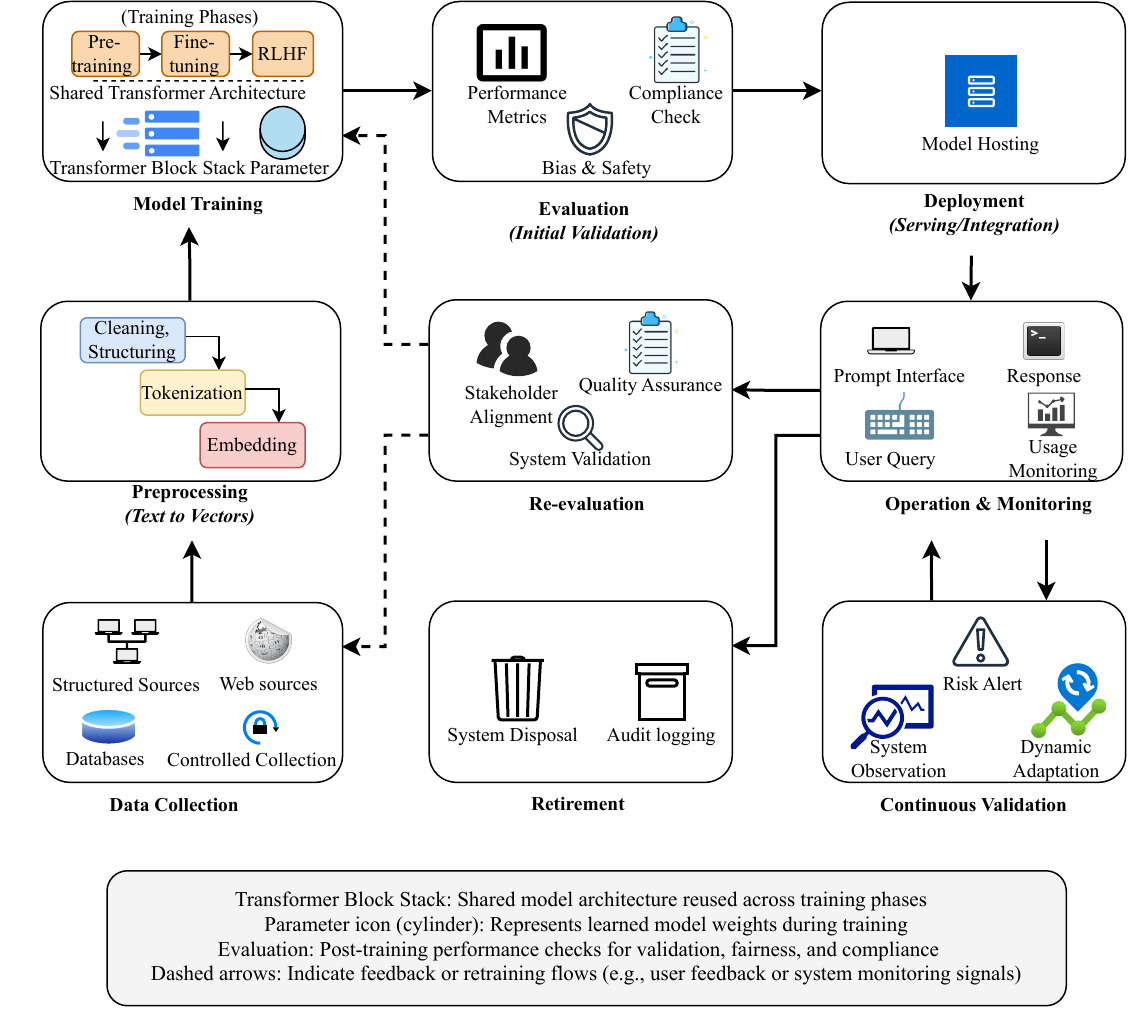}
\caption{LLM System Lifecycle Architecture (adapted from ISO/IEC 5338:2023)}
\label{fig:llm-architecture-updated}
\end{figure*}

an individual to supervise and ensure adherence to regulatory requirements \cite{PIPEDA}. In the context of LLM-based applications, the implementation of accountability can involve the establishment of responsible AI (RAI) and risk management frameworks \cite{xia2024towards}, which support the responsible development, deployment, and oversight of AI solutions. In addition, organizations can conduct periodic Data Protection Impact Assessments (DPIAs) to identify, evaluate, and mitigate emerging risks \cite{kloza2019towards,CNIL2024}. Technical measures can also support accountability through the maintenance of detailed audit logs, documenting data sources, consent events, and model updates.

The next section maps these principles to the lifecycle of an LLM-based system. In addition, it provides a detailed explanation of technical design patterns and organizational controls mentioned in this section and additional ones, illustrating how they can be operationalized in practice across the lifecycle stages of the LLM-based system.

\section{Mapping Regulatory Principles to LLM Architecture} \label{sec:mapping-principles-llm}
In this section, we present our proposed framework for mapping the regulatory principles summarized in Table~\ref{tab:privacy-principles-summary} to four stages of the lifecycle of the LLM-based system: \textit{data collection, model training, operation and monitoring, and continuous validation}. For each principle, we discuss corresponding design, technical, and organizational controls. Before examining how these regulatory principles apply at each stage, we reiterate the challenges of extending principles originally developed for traditional, deterministic information systems to the probabilistic and dynamic behaviour of LLM-based applications.  

\subsection{Challenges in Applying Privacy Regulations to LLM-based Applications}
\label{challenges-applying-law-to-llms}
Ensuring compliance with privacy regulations is particularly challenging for LLM-based applications. Previous research has emphasized that unlike traditional systems such as databases, which store information in discrete records, LLMs are probabilistic and encode information patterns within high-dimensional parameters \cite{feretzakis2025gdpr}. In trained LLMs, the data, including personal information, become embedded in the model parameters, making it difficult to isolate or remove specific information, which makes it particularly challenging to uphold several GDPR rights, including the Right of Access, Right to Rectification and the Right to Erasure or the “Right to Be Forgotten” \cite{feretzakis2025gdpr}. Given that previous research has identified risks associated with data memorization \cite{carlini2022quantifying}, which might later lead to disclosure of such data during inference, it would be critical to allow individuals to access, correct, or request deletion of their data from model parameters. However, as discussed in \cite{feretzakis2025gdpr}, this is currently difficult to achieve in practice, and requires complex technical mechanisms such as machine unlearning \cite{bourtoule2021machine} among other approaches, which currently have some limitations \cite{blanco2025digital}. 

Furthermore, ensuring meaningful consent across the different stages of the lifecycle of LLM-based applications remains a challenge. Children, in their interactions with LLM-based applications, can disclose personal details that services were neither expected nor intended to collect \cite{zhao2019make, irwin2021children}. Although it is possible to request renewed or dynamic consent when new categories of data are processed, in practice, this can be difficult to operationalize, particularly when consent must be managed separately for different data types, processing reasons, and downstream system uses \cite{edpb2020consent}. Consent obtained at a registration may, therefore, not adequately cover the disclosures children make during subsequent interactions \cite{edpb2020consent, wachter2020consent}. To help mitigate this risk, researchers have proposed mechanisms that detect and reformulate sensitive or out‑of‑context information in user inputs before they reach the core system, effectively functioning as input filters or warnings to reduce unnecessary disclosure of private information and protect user privacy during conversational interactions with LLM-based applications \cite{ngong2025protecting}. Nonetheless, there is no guarantee that the system will not capture personal details in conversation histories and later incorporate them into model updates. 

These challenges emphasize persistent gaps when applying existing regulatory principles to LLM-based applications for children. In the proposed framework, we aim to translate regulatory principles into technical controls. At the same time, we acknowledge that, in some instances, adherence to legal requirements and data protection may remain partial.

\subsection{Lifecycle Mapping}
To ensure that privacy protections are proactively embedded into the design of LLM-based applications for children, in accordance with PbD approach, we mapped the regulatory principles and their respective controls to the lifecycle of LLM-based applications. Table~\ref{tab:lifecycle-mapping} illustrates our complete framework: the left column presents the four stages of the LLM-based system lifecycle, the centre column maps specific privacy principles identified in the three regulations, and the right column lists the proposed risk mitigation measures and compliance controls. The next sections discuss each of these in detail. 

\subsubsection{Data Collection}
Data Collection is the foundational phase in the lifecycle of LLM-based applications. During this phase, systems gather raw data, including digital content, books, and code repositories. Data collection takes place before the Model Training phase and may also occur later during the Operation and Monitoring phase. The data collected during the Operation and Monitoring phase can then be used for domain adaptation or model refinement. The entire system lifecycle is presented in Figure~\ref{fig:llm-architecture-updated}, which is adapted from ISO/IEC 5338:2023 \cite{iso2023llm}. Because systems can ingest children’s personal and behavioural data in these corpora, the Data Collection phase must comply with the following regulatory principles: \textit{Data Minimization}, \textit{Purpose Limitation}, and \textit{Meaningful Consent Mechanisms}.

\begin{table*}[ht]
\centering
\caption{Mapping Privacy Principles and Controls Across the LLM Lifecycle}
\label{tab:lifecycle-mapping}
\centering
\footnotesize
\setlength{\tabcolsep}{8pt}
\renewcommand{\arraystretch}{1.15}

\begin{tabular}{
    L{3.0cm}
    L{4.0cm}
    L{9.0cm}
}
\toprule
\textbf{LLM Lifecycle Stage} &
\textbf{Relevant Regulatory Principles} &
\textbf{Technical and Organizational Controls} \\
\midrule

Data Collection &
Data Minimization; Purpose Limitation; Meaningful Consent Mechanisms &
Input filtering to remove unnecessary or sensitive data \cite{ngong2025protecting}; limited collection of identifiers and metadata; purpose-subject data tagging \cite{tokas2020formal}; scoped access controls \cite{spiekermann2009engineering}; verifiable parental consent; age‑appropriate consent interfaces and parental dashboards. \\
\rowcolor{gray!10}

Model Training &
Data Minimization; Security by Design; Accountability &
Task-specific fine-tuning with minimal datasets \cite{staab2025data}; PII removal or anonymization \cite{pasch2025balancing,kuo2025proactive}; encrypted data storage and access controls \cite{ashouri2023privacy, ashouri2024revocable}; validation against data poisoning \cite{das2025security}; differential privacy \cite{elabd2025dynamic}; gradient-based pruning to mitigate backdoors \cite{chapagain2025pruning}; data provenance records \cite{sultan2021ensuring}; DPIAs \cite{kloza2019towards}; Datasheets \cite{gebru2021datasheets} and Model Cards. \\

Operation and Monitoring &
Meaningful Consent; Transparency and Explainability; Purpose Limitation; Security by Design &
Real-time input filtering \cite{ngong2025protecting}; parental dashboards and consent controls \cite{tokas2020formal}; age-appropriate explanations \cite{milkaite2020child, dangol2025children}; purpose-restricted data use \cite{tokas2020formal}; ephemeral session memory \cite{zhang2026burn}; default exclusion of interaction data from training; detection of prompt injection, jailbreaks, unsafe outputs \cite{das2025security}. \\
\rowcolor{gray!10}

Continuous Validation &
Accountability; User Rights; Meaningful Consent Mechanisms; Security by Design &
Periodic audits and DPIAs \cite{raji2022outsider}; logging of system and consent changes; clear interfaces to access, modify or delete data \cite{schaub2015design}; revalidation of parental consent; adversarial testing \cite{liu2023prompt}; integrity checks and model audits \cite{raji2022outsider}. \\
\bottomrule
\end{tabular}

\vspace{0.3em}
\footnotesize\textit{Note: Left Column – LLM Lifecycle Stage; Middle Column – Regulatory Principles; Right Column – Technical and Organizational Controls.}
\end{table*}

\textit{Data Minimization} requires applications to limit data collection to only what is necessary for their intended use \cite{gdpr2016,PIPEDA}. GDPR Article 5(1)(c), PIPEDA Principle 4 along with implicit coverage in COPPA, all prohibit gathering more information than is reasonably required to enable participation in the online activity or service. To apply this principle during the Data Collection phase, designers and developers can use strategies such as input-filtering through techniques like those deployed in \cite{ngong2025protecting} whereby user inputs are filtered and re-formulated to remove unrelated and potentially sensitive information.  This limits the number of data fields collected. For example, systems should not collect identifiable information, such as full names, birth dates, or contact information, unless it is collected on lawful basis \cite{gdpr2016}. As discussed in earlier sections, a key vulnerability of LLMs is their ability to memorize strings of training data, which may be reproduced verbatim during inference time \cite{carlini2022quantifying}. As such, the main goal of data collection phase with regards to privacy is to ensure that private and sensitive data do not make it to the training corpora. Furthermore, systems should not automatically save background data (such as cookies, device IDs, or metadata) or interaction data for model enhancement by default, and should only use these data when their use is clearly communicated and supported by valid consent. 

\textit{Purpose Limitation} states that the system should collect data only for specific, legitimate purposes, and should not re-purpose them without obtaining additional consent. This principle is codified in GDPR Article 5(1)(b) and reflected in PIPEDA Principle 5. COPPA Rule Sections 312.4 and 312.5 require operators to clarify purposes and obtain new verifiable consent if those purposes change. For example, when a system collects data to customize tutoring responses, it must not use that information for behavioural advertising or analytics unless there is a valid legal basis or renewed consent is obtained. To support this principle, privacy engineering frameworks suggest various technical controls, including purpose-specific data tagging \cite{tokas2020formal}, scoped access controls \cite{spiekermann2009engineering}, and provenance-enabled access control to preserve the initial collection purpose throughout processing \cite{sultan2021ensuring}. These methods reduce the risk of unintended data reuse and strengthen compliance with regulatory requirements. 

\textit{Meaningful Consent Mechanisms} require that, before data collection or processing occurs, the system obtains verifiable parental or guardian consent, as required by Section 312.5 of the COPPA Rule, Articles 6 and 8 of the GDPR, and Principle 3 of PIPEDA. While the legal responsibility rests with parents or guardians to provide consent, children should still receive age-appropriate explanations of how the system processes data to ensure transparency and respect for their developing autonomy. Consent interfaces should, therefore, be child-friendly; they should use simplified language, visualizations or video explanations to convey how the system will use their data \cite{milkaite2020child}. The system must also include verification methods to ensure that consenting adults are of legal age (e.g., knowledge-based questions, verification through a bank account, or a digital signature) \cite{coppa2020guidelines}. Furthermore, consent should remain dynamic \cite{tokas2020formal,khalid2023enhancing}. Dynamic consent allows parents or guardians to review, update, or revoke consent. The formal framework for consent management presented in \cite{tokas2020formal} demonstrates how dynamic consent can be achieved in practice. Their framework enables users (e.g. parents or guardians) to access and view current privacy settings in real-time through a user-friendly interface, like a privacy dashboard, and update these settings when their preferences change.

\subsubsection{Model Training} \label{subsec:model-training-phase}
The Model Training phase turns raw data into learned representations. This phase presents additional privacy risks if earlier safeguards fail and personal or sensitive information, or data that can indirectly identify someone, enters model parameters. To reduce these risks, developers should follow the principles of \textit{Data Minimization}, \textit{Security by Design}, and \textit{Accountability} to ensure that sensitive or personal information does not enter the training process.

\textit{Data Minimization} requires using only the smallest and most relevant subset of data required \cite{staab2025data}. Service providers can align their practices with legal requirements by fine-tuning a foundational model on datasets directly applicable to the task. For example, a math tutoring model should rely on math-related educational content. During further refinement with user interactions, developers must include only the interactions for which explicit consent was given by parents to allow use in model refinement. These data should have PII and other sensitive content removed. Techniques for achieving this include input-filtering, anonymization, and PII reduction methods \cite{ngong2025protecting,pasch2025balancing,kuo2025proactive}. 

\textit{Security by Design} requires safeguards at every stage of model training. Article 32 of GDPR, Principle 7 of PIPEDA, and Section 312.8 of the COPPA Rule collectively mandate the implementation of appropriate technical and organizational measures to protect personal data. During model training, datasets should be encrypted in both transit and rest, access to training data must be restricted through access control mechanisms, and data sources should be thoroughly analyzed and validated to mitigate data poisoning attacks, in which adversaries deliberately manipulate training data to compromise model decision-making \cite{das2025security}. Security teams must also protect training pipelines from privacy attacks, including gradient leakage attacks, in which adversaries exploit shared gradients to reconstruct sensitive training data. A commonly adopted approach to mitigating such risks is the use of differential privacy \cite{das2025security}, which introduces calibrated noise into gradients or parameter updates during training \cite{elabd2025dynamic}. Prior work has shown that differential privacy can provide privacy guarantees while maintaining an acceptable model utility \cite{abadi2016deep}. In addition, security teams must also protect trained models from backdoor attacks, in which adversaries insert hidden tokens to manipulate a model to behave in a particular way, for example, generating malicious output when a trigger token is invoked \cite{das2025security}. Recent studies suggest techniques such as gradient-based pruning,  which removes model components that show low sensitivity to loss gradients which might potentially be backdoor triggers \cite{chapagain2025pruning}.        

\textit{Accountability} helps organizations demonstrate compliance with data protection laws and uphold ethical principles. Article 5(2) of GDPR clearly requires accountability, similar to the obligations outlined in PIPEDA Principle 1. To meet these principles, stakeholders should implement responsible AI and risk management frameworks \cite{xia2024towards}, maintain data provenance records \cite{sultan2021ensuring}, verify consent for data related to children \cite{tokas2020formal}, and apply privacy and security protections against attacks \cite{das2025security}. In addition, DPIAs can help identify high-risk data categories \cite{kloza2019towards}. Tools such as Datasheets for Datasets \cite{gebru2021datasheets} and Model Cards  can explain how the system handles data and uses models, improving transparency, trust, and ethical practices, particularly for applications designed for children.

\subsubsection{Operation and Monitoring} \label{subsec:operation-monitoring-phase}
During the operation and monitoring phase, LLM-based applications interact directly with users to provide real-time output. This phase introduces unique privacy and security challenges, particularly for children. Because interactions between children and LLM-based applications are live and dynamic, they involve ongoing data exchange and adaptive system responses, during which children can unintentionally reveal sensitive or personal information \cite{ngong2025protecting}. Such data may be stored in conversation histories and may be used for future model refinement or retraining \cite{zanella2020analyzing}. Furthermore, in some jurisdictions, temporary storage for purposes such as safety monitoring or compliance with law enforcement requests is permitted, which can potentially increase the risk of unauthorized insider access. Another set of risks relates to information sharing activities with third parties. To reduce these risks, developers should apply the principles of \textit{Meaningful Consent}, \textit{Transparency and Explainability}, \textit{Purpose Limitation}, and \textit{Security by Design}. These safeguards should be complemented by continuous supervision and safety monitoring to ensure responses remain appropriate for children and to detect potential disclosures of distress or self-harm. There is a growing area of research in this direction \cite{holmes2025applications}. 

\textit{Transparency and Explainability} help build trust and support informed oversight in interactions between parents or guardians, their children, and LLM-based applications. Articles 12–13 of the GDPR and Section 312.4 of the COPPA Rule establish transparency requirements, calling for clear and accessible information about how personal data are processed, with particular attention to age-appropriate communication. The system should therefore provide simple explanations on what kind of data it will collect during interactions with children, how these data are used, stored, and shared. In LLM-based applications, transparency might also involve explanations on how outputs are generated and how user inputs influence those outputs. Service providers can implement parental dashboards with clear and complete information on data flows to facilitate informed decision-making. In addition, service providers can implement visualizations and video-based explanations, as well as reduce the reliance on dense textual disclosures to support meaningful transparency, especially when communicating with children \cite{milkaite2020child}. Furthermore, system developers should involve children in the design of transparency mechanisms. Participatory and co-design approaches that incorporate children’s preferences and perspectives can improve visibility into data processing using AI \cite{milkaite2020child}. Recent work on children’s mental models of AI further suggests that it is useful to address common misconceptions and provide meaningful explanations of how AI decisions are made \cite{dangol2025children}.

\textit{Meaningful Consent} helps ensure parents or guardians have control over what happens to their children's data once they start interacting with an LLM-based application. According to COPPA Rule 312.5 (Parental Consent), and GDPR Articles 6 and 8, parental or guardian consent must be verifiable. This means, service providers should maintain logs of parental or guardian consent. Consent should also be viewed as an ongoing process, reviewed whenever data use or system functionality changes, or when parents or guardians wish to make changes to what data about their children they want to share. As discussed above, this can be supported through a dynamic consent management framework, as proposed in \cite{tokas2020formal}. Parental or guardian dashboards can help families understand what information is collected, adjust permissions, or withdraw consent \cite{tokas2020formal}. This enables user autonomy while supporting compliance with regulatory requirements \cite{cifar2024responsible,yu2025exploring}. 

\textit{Purpose Limitation} ensures that user input during the Operation and Monitoring phase is used only for the immediate conversational context. This principle blocks secondary uses, such as behavioural profiling, analytics, or fine-tuning, unless there is a valid legal basis or explicit renewed consent. To implement these legal requirements, service providers must obtain parental or guardian consent discussed earlier. The purpose and subject tagging can then be used to ensure the data collected from children is only used for purposes for which parental or guardian consent was provided \cite{tokas2020formal}. To avoid data leakage during interactions, systems can use ephemeral session memory designs to delete conversational context and temporary embeddings after each use \cite{zhang2026burn}. Furthermore, LLM-based applications used by children should not collect user input for model training or sharing data with third parties, by default; such collection should be enabled only through parental or guardian controls. The only exception is when the sharing with third parties is necessary to provide the service \cite{coppa2020guidelines}.    

\textit{Security by Design} is important during live operations to protect against threats such as prompt injections, where malicious input bypasses LLM safety filters and manipulates model output, for example, to reveal sensitive information \cite{das2025security}; jailbreaks, which manipulates the application's safety mechanisms to elicit prohibited responses \cite{das2025security}; and inference leakage, where the model inadvertently reveals sensitive or training-related information \cite{duan2024membership}. This principle is required under GDPR Article 32 and COPPA Rule Section 312.8. To maintain security during the Operation and Monitoring phase, security teams can implement input filtering to block harmful and adversarial prompts and perform runtime output filtering for unsafe content \cite{das2025security}. Earlier measures to reduce PII in training data can prevent it from being revealed during inference \cite{das2025security}. 

\subsubsection{Continuous Validation}
\label{subsec:continuous-validation-phase} 
The Continuous Validation phase involves the ongoing oversight, monitoring, and governance of an LLM-based application after its deployment \cite{myllyaho2021systematic}. This phase ensures that the system remains accurate and reliable, secure, and maintains ethical standards \cite{IEEE2024_LLM_Lifecycle}. Unlike the Operation and Monitoring phase, which focuses on real-time user interactions, the Continuous Validation phase addresses long-term system integrity, mitigates biases, incorporates new knowledge and responds to emerging challenges \cite{IEEE2024_LLM_Lifecycle}. This phase also involves verification of compliance with privacy and safety standards. Four principles are important during this phase: \textit{Accountability}, \textit{User Rights}, \textit{Meaningful Consent Mechanisms}, and \textit{Security by Design}.

\textit{Accountability} is necessary when systems undergo updates, re-evaluations, and ongoing monitoring. Articles 5(2) and 24 of the GDPR, and Principle 1 of PIPEDA require organizations to demonstrate their compliance with regulatory requirements. As discussed before, effective data and AI governance program as well as risk identification and mitigation can support organizations in achieving transparent, ethical, and auditable AI systems that align with regulatory requirements \cite{xia2024towards}. Additionally, DPIAs, along with regular audits and evaluations, help identify issues that arise from feature updates and changes \cite{raji2022outsider}. Because iterative updates can introduce new risks and affect the way user rights are provided, maintaining traceability throughout the lifecycle is a key challenge. To reinforce transparency and trust, organizations should provide clear and accessible channels for regulatory inquiries, user complaints, and rights requests as required under GDPR Article 12 and supported by GDPR Recital 39.

\textit{User Rights} must be actively enforced during the continuous validation phase, as mandated by Section 312.6 of the COPPA Rule, Articles 15-17 of the GDPR, and Principle 9 of PIPEDA. These rights include accessing, correcting, or requesting deletion of personal data, especially critical for child-specific applications. To protect these rights, service providers can provide clear notices, pathways and interfaces support user engagement and understanding \cite{schaub2015design}. Such interfaces can allow parents or guardians to exercise these rights, such as requesting deletion of their child’s data, including conversation records, voice recordings, images, and other media, in a clear and actionable manner. Additionally, service providers can implement automated request workflows that log and time-stamp actions, creating a verifiable audit trail that supports regulatory compliance and strengthens trust \cite{samavi2018publishing}.

\textit{Meaningful Consent Mechanisms} must extend beyond initial onboarding to maintain compliance with COPPA Rule Section 312.5, GDPR Articles 6 and 8, and PIPEDA Principle 3. Consent must remain verifiable and informed, allowing parents or legal guardians to authorize, monitor, and revoke data collection or processing at any time. During continuous validation, this involves periodic revalidation, prompting guardians to review and reaffirm consent whenever new features, data uses, or policy changes arise. 

\textit{Security by Design}
is in accordance with COPPA Rule Section 312.8, GDPR Article 32, and PIPEDA Principle 7, which require controls to protect personal information from unauthorized access and modification, or inference leakage. For LLM-based applications designed for children, this involves implementing input and output filtering to prevent harmful or inappropriate content displayed to children. Adversarial testing can help identify and protect from prompt injection \cite{liu2023prompt} and jailbreak attacks \cite{das2025security}. Additional measures, such as strong authentication, encrypted storage and transmission, and hash-based model integrity checks, enhance data protection and auditability \cite{shevlane2022structured}. Privacy-preserving techniques, including differential privacy, further protect sensitive data from being reconstructed in system responses \cite{liu2025artificial,miranda2024preserving}. Periodic model audits and assessments should detect behavioural changes or data drift caused by authorized and unauthorized updates, providing ongoing verification of technical integrity and regulatory compliance \cite{raji2022outsider}.

\subsection{Anthropomorphic and Nudging Risks}
\label{subsec:special-considerations-for-children} 
In addition to the privacy risks associated with LLM-based applications previously discussed, it is important to highlight a new risk: that these applications may present themselves as human beings. This can potentially influence children to disclose more personal information than they would normally share. Therefore, it is crucial to design LLM-based applications in a way that does not unduly influence children.

The United Nations Convention on the Rights of the Child (UNCRC) \cite{uncrc1989}, the \textit{Age Appropriate Design Code} by the UK Information Commissioner's Office (ICO) \cite{ico2020aadc}, and previous work provide valuable guidance for achieving this objective. Regarding children's rights, Article 17 of the UNCRC emphasizes a child's right to access information that promotes their social, spiritual, and moral well-being, while also requiring the establishment of appropriate guidelines to protect children from harmful information \cite{uncrc1989}. Based on this definition, children have the right to know how service providers collect their data. 

Similarly, the AADC Transparency standard encourages developers to be open and transparent about their services, including how children's personal data is collected, used, and shared, and to communicate this information in a manner appropriate for their age. The guidance further notes that transparency is linked to the principle of fairness: when service providers are unclear, opaque, or dishonest about how their services operate or how data is collected, the resulting data processing and use are likely to be unfair \cite{ico2020aadc}. 

Anthropomorphic design may contravene both the right to information under the UNCRC and the AADC transparency guidance, because children may not recognize that they are disclosing personal information when interacting with systems that present themselves as friendly or empathetic. Some researchers have therefore expressed concern that anthropomorphic design can function as a manipulative data-elicitation practice \cite{kurian2025developmentally, weidinger2021ethical}. These guidelines, taken together, suggest that service providers should be transparent about their data collection practices. Anthropomorphic effects raise concerns because they may (1) unduly influence children's disclosure behaviours and (2) result in data collection and use that might not be sufficiently transparent.   

Reducing anthropomorphic effects is therefore a proactive measure and a design objective to protect privacy. Current recommendations suggest that children should be aware that they are communicating with an information system, not a human being \cite{weidinger2021ethical}. For this, the systems need to explicitly identify itself as an AI system \cite{weidinger2021ethical, kurian2025developmentally}. Reminders can also be added throughout interactions. For instance, systems may explicitly inform children with messages such as, “I am a machine and do not possess an understanding of human emotions” \cite{kurian2025developmentally}. Additional measures suggested in existing research to reduce anthropomorphic effects might involve preventive experiments in which developers identify whether certain agent dispositions or responses lead to more information disclosures and remove them before the system is released to actual users \cite{akbulut2024all}. 
   
Nudging has been a recognized phenomenon in information systems, and, in fact, AADC explicitly recognizes and calls for avoiding nudging techniques for a child-centred AI \cite{ico2020aadc}. Nudging, as discussed in Section \ref{sec:related-work}, refers to the design of online services that encourage children to provide more personal information than they would otherwise volunteer \cite{ico2020aadc}. In the case of interactive agents, nudging can take the form of follow-up prompts such as “Tell me more” or “What do you feel like right now?”, which can lead to more information disclosure. AADC standard guides against the use of any nudge techniques that undermine privacy. In addition, it encourages the use of pro-privacy nudges where appropriate and nudges that promote their health and well-being \cite{ico2020aadc}. In practical cases involving LLM-based technologies, this can look like removing follow-up questions from system design, setting reminders that the information provided to systems might be accessed and read, and encouraging speaking to parents or guardians when certain information is shared.        

\subsection{Case Study: Educational LLM Tutor for Children} \label{subsec:educational-tutor}
\begin{figure*}[!t]
\centering
\includegraphics[width=0.75\textwidth]{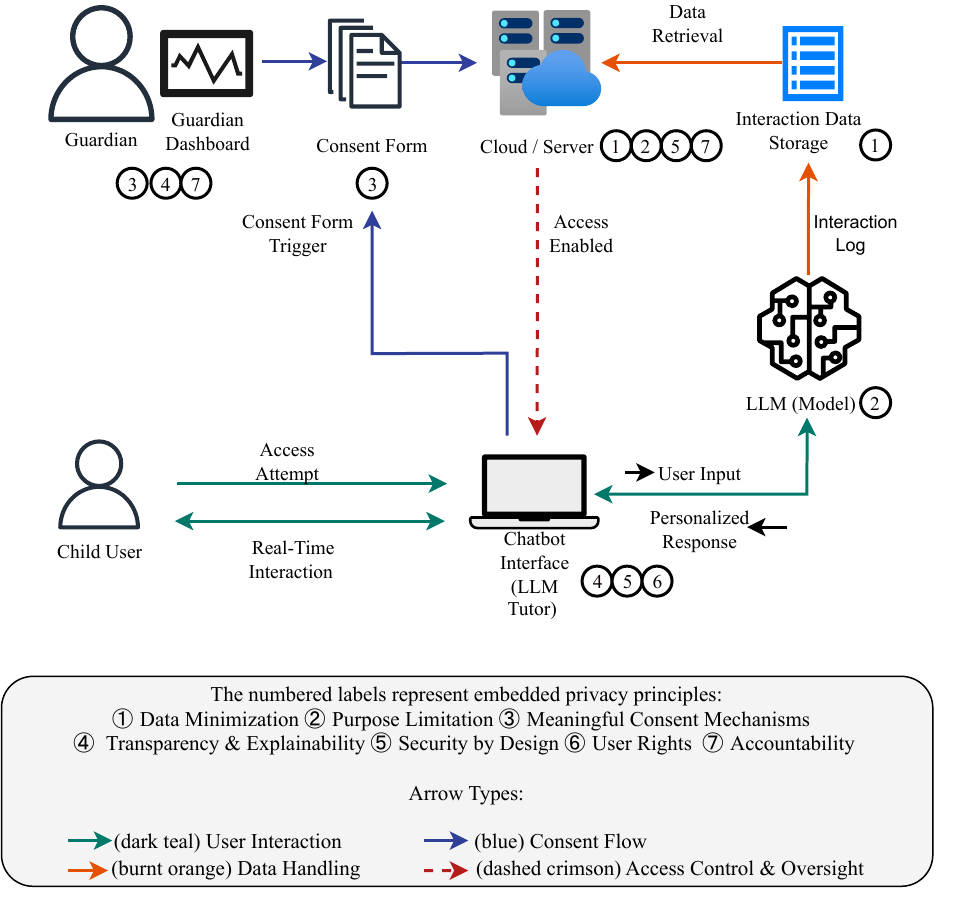}
\caption{Educational LLM Tutor Architecture Aligned to Lifecycle Stages and Privacy Principles}
\label{fig:llm-tutor-architecture_new}
\end{figure*}

We present an LLM-based educational tutor for children under 13 to demonstrate how we can operationalize the proposed framework in a real-world application deployment (Figure \ref{fig:llm-tutor-architecture_new}). The application acts as a personalized homework assistant for math, language, arts, and science. Children interact through a conversational chatbot interface, powered by an LLM on the background. 

\textit{Data Collection} Principles include \textit{Data Minimization}, \textit{Purpose Limitation}, and \textit{Meaningful Consent}. 
Parents and guardians initiate the onboarding with verifiable parental consent (e.g., a microcharge on a credit card or an equivalent method such as ID verification), providing age-appropriate disclosures and a consent dashboard that supports granular choices made that can be revisited at a later date. The system minimizes and pseudonymizes inputs (input filtering), blocking off background identifiers such as cookies, device IDs, and precise location by default and only activating them with explicit, specific consent. Purpose tags are also attached to data at ingress, where training uses opt-in by default for children's data.

\textit{Model Training.}
Principles include \textit{Data Minimization}, \textit{Security by Design}, and \textit{Accountability}.
The system filters corpora gathered from child-tutor interactions and maintains exclusion lists for direct identifiers (e.g., names, addresses, etc.). The system uses differential privacy and strict access controls for any child-related fine-tuning \cite{dwork2014algorithmic}. The documentation records all training data lineage, lawful basis, and consent artifacts (in datasheets and model cards) to support audits and unlearning requests.

\textit{Operation and Monitoring.}
Principles include \textit{Transparency/Explainability}, \textit{Purpose Limitation}, \textit{Security by Design}, and \textit{User Rights}. 
The chatbot identifies itself as an automated system, avoids human-like cues, and uses neutral language in its responses. Its dialogue policies are designed to prevent prompts that might elicit unnecessary personal disclosures. By default, its session memory is ephemeral, and conversation logs are minimized, encrypted, and retained only for explicitly stated safety or quality assurance purposes, under settings controlled by parents or guardians.

Figure~\ref{fig:llm-tutor-architecture_new} illustrates the architecture of the LLM-based educational tutor, highlighting the alignment between the phases of the system lifecycle and the principles of privacy. The chatbot communicates with a child in an age-appropriate language. A parent or guardian can review the summary of the interactions. Rather than exposing full transcripts, the dashboard for parents or guardians presents aggregated insights such as discussion topics and safety alerts, providing oversight without undermining the child’s autonomy or trust \cite{ico2020aadc}. 
Model retraining on children's interactions is disabled by default and can only be enabled with an explicit parental or guardian consent. Runtime controls include input sanitization, adversarial prompt detection, output filtering, and secure data transmission and storage to prevent unauthorized access.

\textit{Continuous Validation.}  The system performs ongoing monitoring and evaluation of model behaviour, consent compliance, and privacy metrics, allowing detection of anomalies, drift, or privacy violations and supporting timely updates or interventions to maintain child safety and regulatory compliance.

\section{Discussion} \label{sec:discussion}

Given the rapid expansion of AI-based systems and children’s increasing engagement with them, governance mechanisms for child-facing AI systems must anticipate risks rather than respond to harms after they materialize. Our review shows that children may face multiple privacy risks that accumulate across the lifecycle of LLM-based applications. These include technical vulnerabilities, governance gaps, and design choices that may disproportionately affect children. Embedding privacy as a foundational design requirement can help mitigate these risks and ensure that AI systems are appropriate, lawful, and protective of children’s rights.

To embed privacy protections into the design of LLM-based applications for children, we proposed an integrated framework based on the PbD approaches, that translates regulatory privacy principles from the GDPR, COPPA, and PIPEDA into actionable technical and organizational controls across the lifecycle of these systems. The framework maps regulatory requirements to the lifecycle stages of LLM-based applications, including Data Collection, Model Training, Operation and Monitoring, and Continuous Validation. This mapping identifies where privacy risks emerge and how designers and developers can mitigate them through design-time and operational controls informed by recent academic literature on privacy, security, and child-specific system design practices.

In the remainder of this section, we examine how our proposed framework operationalizes PbD principles in children’s LLM-based applications and discuss the technical, regulatory, and practical implications of adopting a lifecycle-oriented privacy protection approach. We highlight both the strengths of this approach and the persistent challenges that remain when translating legal norms into deployable systems for child-facing AI.

\subsection{Alignment Between the Proposed Framework and PbD}
Our proposed framework operationalizes the seven PbD principles throughout the lifecycle of LLM-based applications for children: proactive risk management, privacy by default, privacy embedded into design, positive-sum functionality, end-to-end security, visibility and transparency, and respect for user privacy across the LLM lifecycle. Key technical measures include input filtering, subject-purpose-based data tagging, verifiable parental consent, adversarial training, differential privacy, ephemeral session memory, guardian dashboards, secure data transmission, age-appropriate explanations, and avoidance of human-like cue and nudging, among others.

Specifically, our framework aligns practices with the first principle, proactive risk management, by implementing input filtering to minimize data collection and by proposing subject-and purpose-based tagging, which reduces the risk of over-collection and unauthorized uses. The framework further supports privacy as a default principle by prohibiting the collection of user interactions for model training by default, unless parents or guardians explicitly enable it or provide consent. It supports positive-sum functionality by balancing privacy protections with system utility. Specifically, the framework shows how systems can still provide educational benefits without requiring unnecessary data collection, such as names or ages. The framework aligns with the principle of end-to-end security by incorporating adversarial training, privacy-preserving techniques such as differential privacy, ephemeral session memory, and secure data storage and transmission, among other measures. The framework enables visibility and transparency through parental and guardian dashboards, age-appropriate explanations, and auditable system logs that enhance observability and accountability. Finally, the framework supports respect for user privacy throughout the LLM lifecycle through verifiable parental or guardian consent mechanisms, dynamic consent, and ongoing parental oversight. The framework also addresses child-specific considerations, such as age-appropriate explanations and minimizing nudging or human-like cues, further supporting ethical and legally compliant LLM deployment.

\subsection{Limitations}

Although this work demonstrates that PbD principles can be operationalized through regulatory and technical controls in LLM-based applications for children, significant challenges and gaps remain in the four key stages of the LLM lifecycle. These must be addressed to move from privacy-aware prototypes to scalable, legally compliant, and ethically robust deployments.

\subsection{Challenges in Implementing PbD Principles Across LLM-lifecycle}
\label{subsec:challenges-pbd-llm}

\textit{Data Collection.} Even with minimization policies and parental consent workflows in place, children may still disclose personal or sensitive details through open-ended interactions given they might not fully understand online privacy risks \cite{zhao2019make}. Additionally, it has been documented that children often access LLM-based applications to seek emotional support and companionship \cite{yu2025exploring}, which can reveal a lot of private details. Current input filtering and entity redaction techniques provide partial protections by removing explicit identifiers such as names, locations or phone numbers \cite{ngong2025protecting}, however children may use indirect phrasing, speak of personal events without identifiers, which can still reveal details about them, their lifestyles, or associations. For example, a child can mention “the park near grandma’s house” or “my school trip tomorrow,” which cannot always be captured by rule-based or keyword-driven filters. These limitations highlight the residual risk of identifiable data entering the system despite compliance mechanisms. To address this, future work should focus on adaptive filtering methods that combine natural language understanding, context-awareness, and child-specific linguistic patterns to detect disclosures. Importantly, such approaches must strike a balance between privacy protection and usability, ensuring that children are not discouraged from meaningful engagement with educational or assistive applications.

\textit{Model Training.} A persistent challenge is to mitigate model memorization, where rare or sensitive inputs, such as a child’s name, address, or personal story, become embedded in model parameters and risk being revealed during inference. This risk is heightened in child-specific applications because children may inadvertently share uniquely identifiable details in ways adults do not. While techniques such as differential privacy offer partial protection, empirical studies show that LLMs can still reproduce memorized phrases when subjected to targeted prompting or extraction attacks \cite{kassem2023preserving,zhou2024quantifying}. Current defenses often struggle to balance utility with privacy, as strong noise addition can degrade the accuracy and general usability of the model \cite{abadi2016deep, dwork2014algorithmic}. Emerging approaches, such as scalable machine unlearning methods \cite{blanco2025digital} that selectively erase sensitive records without requiring full retraining, offer promising directions. Developing such strategies would directly reduce the long-term risks of exposing children’s personal information, while maintaining model quality for safe educational and developmental use.

\textit{Operation and Monitoring.} Providing meaningful transparency to children, parents and guardians remains difficult due to the probabilistic and opaque nature of LLMs. Simply logging responses or publishing model cards is insufficient, as these artifacts often do not capture how sensitive information is processed in real time \cite{mitchell2019model, raji2022outsider, samavi2018publishing}. The interfaces must therefore evolve to translate complex reasoning processes into age-appropriate explanations for children \cite{dangol2025children} and actionable oversight insights for parents and guardians. However, achieving this balance requires advancing explainability methods beyond post-hoc interpretations towards accountability dashboards that visualize various risks associated with a particular LLM-based tool \cite{yu2025exploring}. Dynamic consent management further complicates this challenge: while static consent forms exist, mechanisms for guardians to update or revoke permissions in real time are underdeveloped, particularly for unstructured, conversational input streams, which characterize current AI technologies \cite{skulmowski2025informed}. Consent, therefore, must be viewed as a living document \cite{skulmowski2025informed}. Research into adaptive monitoring frameworks, coupled with lightweight privacy-preserving audit trails, is critical to operationalizing transparency and consent in practice \cite{samavi2018publishing, pearson2013accountability}. Finally, although ephemeral session memory can reduce privacy risk, it may conflict with safety, abuse prevention, and debugging needs, and therefore systems must clearly specify what is retained (if anything), for how long, and under what access model and purpose constraints.

\textit{Continuous Validation.} Maintaining privacy and safety protections after deployment is challenging due to both evolving threats and jurisdictional differences \cite{iso2023llm, das2025security}. Protective mechanisms implemented at launch can degrade over time if not reinforced through periodic testing, red teaming, and model audits \cite{qi2024evaluating, raji2022outsider}. Moreover, cross-border use of LLM-based applications for children demands governance architectures that harmonize COPPA, GDPR, and PIPEDA obligations while adapting to local enforcement nuances \cite{neel2023privacy,ico2020aadc}.

Finally, current engineering frameworks for AI are not well-suited to the developmental and cognitive needs of children. There is a critical need for children-specific interface patterns, consent workflows, and privacy risk models that integrate both technical and developmental protections \cite{unicef2021ai}. Without such adaptations, PbD risks remaining a high-level aspiration rather than a fully operationalized standard in LLM-based ecosystems for children.

\subsection{Limitations of the Proposed Framework} \label{subsec:limitations}

Although the proposed framework demonstrates how privacy and regulatory principles can be mapped onto the architecture of a child-specific LLM tutor, several limitations remain. First, the framework operates at the conceptual and architectural level. Translating these into production systems requires significant engineering effort, financial investment, and organizational commitment \cite{diaz2024large, iso2023llm}. In practice, smaller developers or educational institutions may struggle to implement the full set of recommendations, particularly those that require advanced cryptographic methods or dedicated compliance teams. Additionally, smaller developers may lack leverage against large vendors that control the infrastructure.

Second, some of the technical protective mechanisms, such as differential privacy, machine unlearning, and tamper-evident logging, are still areas of active research \cite{ginart2019making, samavi2018publishing, yu2021differentially}. Although promising in theory, their deployment at scale in multibillion-parameter models remains experimental and may introduce trade-offs in accuracy, usability, or system performance. This raises questions about whether current technological maturity is sufficient to meet the high expectations set by regulatory frameworks like GDPR, COPPA, and PIPEDA \cite{feretzakis2025gdpr, egelman2023informing}.

Finally, the legal and social contexts are dynamic. Regulations evolve, cultural expectations of children’s digital rights differ between jurisdictions, and parents can vary in their ability or willingness to engage with dashboards and consent mechanisms \cite{livingstone2021rights, yu2025exploring}. The proposed model cannot fully capture these contextual nuances, and therefore, its effectiveness will depend on ongoing adaptation, stakeholder participation, and iterative design informed by real-world use.

Collectively, these limitations indicate that the proposed framework should be regarded as a foundational basis for ongoing development. It is the collaboration among technologists, educators, regulators, and families, which will be essential to refine and operationalize these principles in practice.

Nonetheless, the proposed framework represents a meaningful step toward designing safer, more secure, and more compliant LLM-based tools for children. Despite the unprecedented exposure to these tools during the formative years and their unique vulnerabilities, children remain underrepresented in discussions of AI risks. Recent policy developments, particularly in the United States and increasingly in the United Kingdom, emphasize age verification as a primary mechanism for protecting children’s online privacy and safety \cite{dpc2021childfundamentals,opcChildrenCode}. While these measures may reduce underage access to certain applications, they risk overemphasizing verification at the expense of broader privacy-preserving mechanisms. Combining age verification with additional measures, such as those proposed in our framework, including data minimization, dynamic consent, security by design, and protections throughout the data lifecycle, offers a more balanced approach. Together, these mechanisms help build trust with parents and guardians, who need assurance that their children’s data is handled responsibly \cite{long2023trust,pearson2013accountability}, and support children in exploring and interacting with technology while minimizing risks to their privacy.

\section{Conclusion} 
\label{sec:conclusion}

In this paper, we examined how privacy and data protection principles derived from COPPA, GDPR, and PIPEDA can be operationalized in the design and deployment of LLM-based applications for children. Through comparative legal and technical analysis, we articulated seven core regulatory principles, \textit{Data Minimization}, \textit{Purpose Limitation}, \textit{Meaningful Consent Mechanisms}, \textit{Transparency and Explainability}, \textit{Security by Design}, \textit{Accountability}, and \textit{User Rights}, as fundamental for the development of ethically aligned and legally compliant LLM systems \cite{gdpr2016,coppa2020guidelines,PIPEDA}.

We mapped these principles to four stages of the LLM lifecycle: \textit{Data Collection}, \textit{Model Training}, \textit{Operation and Monitoring}, and \textit{Continuous Validation}. Each stage presents unique privacy risks for children, requiring technical and organizational controls to support compliance and protect child data. To address these risks, we proposed a PbD-aligned privacy engineering framework informed by guidance from data protection authorities and children’s rights organizations. Some recommended protective mechanisms include local or ephemeral data processing, session-based memory isolation, real-time input filtering, and guardian-based dashboards to enhance transparency and oversight. Where limited retention is necessary for safety monitoring, abuse prevention, or incident response, retained data should be minimized, time-bounded, access-controlled, and purpose-separated from model training or unrelated analytics. Security-by-design controls in this context are also directly tied to child privacy outcomes, because failures such as prompt injection, jailbreaks, poisoning, or backdoors can lead to disclosure of sensitive child data, bypass of parental controls, or unsafe elicitation of unnecessary personal information. The proposed framework demonstrates how abstract legal obligations can be translated into enforceable technical controls, accessible interfaces, and auditable governance, supporting both regulatory compliance and the cultivation of trust in LLM-based applications for children. Using an educational LLM tutor case study, we illustrated how these principles can be operationalized in real-world contexts. By operationalizing these privacy principles into actionable design practices, this work provides a roadmap for supporting compliance with legal requirements in LLM-based applications for children, along with enhancing safety, privacy, and well-being of children. 
 
While controls mentioned in our framework offer improved privacy protection, persistent vulnerabilities remain, including implicit profiling, model memorization of sensitive disclosures, and insufficient transparency for parents or guardians \cite{mireshghallah2023can,neel2023privacy,carlini2021extracting}. Additionally, the proposed framework remains conceptual. Future research should focus on advancing scalable technical solutions that extend beyond conceptual frameworks. Priority directions include improving robustness against memorization and extraction attacks, developing explainability methods tailored to children and guardians, enabling dynamic consent management through mechanisms such as verifiable credentials or smart contracts \cite{kassem2023preserving,uddagiri2024ethical,kloza2019towards}, and supporting cross-border regulatory interoperability for globally deployed LLM systems.

\label{subsec:technical-interpretation}

\bibliography{1bibfile}

\end{document}